\documentclass[pdflatex,sn-mathphys-num]{sn-jnl}
 
\usepackage{graphicx}%
\usepackage{multirow}%
\usepackage{amsmath,amssymb,amsfonts}%
\usepackage{amsthm}%
\usepackage{mathrsfs}%
\usepackage[title]{appendix}%
\usepackage{xcolor}%
\usepackage{textcomp}%
\usepackage{manyfoot}%
\usepackage{booktabs}%
\usepackage{algorithm}%
\usepackage{algpseudocode}
\usepackage{listings}%
\usepackage{changepage}  
\usepackage{microtype}
\usepackage{comment}

\begin{document}


\title[Article Title]{ChatGPT in Research and Education: Exploring Benefits and Threats}
 \author*[1,4]{\fnm{Abu Saleh Musa} \sur{Miah}}\email{abusalehcse.ru@gmail.com}
 \author[1]{\fnm{Md Mahbubur Rahman } \sur{Tusher}}\email{iiauthor@gmail.com}
 \equalcont{These authors contributed equally to this work.}
 \author[1]{\fnm{Md. Moazzem} \sur{Hossain}}\email{moazzem.cse10@gmail.com}
\equalcont{These authors contributed equally to this work.}
 \author[1]{\fnm{Md Mamun} \sur{Hossain}}\email{mamunsust12@gmail.com}
 \equalcont{These authors contributed equally to this work.}
 \author[2]{\fnm{Md Abdur} \sur{Rahim}}\email{iiiauthor@gmail.com}
 \author[3]{ \fnm{Md Ekramul} \sur{Hamid}}\email{ekram\_hamid@yahoo.com }
 \author[3]{ \fnm{Md. Saiful } \sur{Islam}}\email{msis\_ru@yahoo.com }
 \author*[4]{\fnm{Jungpil} \sur{Shin}}\email{jpshin@u-aizu.ac.jp }
 \affil[1]{\orgdiv{Department of CSE},
 \orgname{Bangladesh Army University of Science and Technology (BAUST)} 
 \orgaddress{\city{Saidpur}, \postcode{5310}, \state{Rangpur}, \country{Bangladesh}}}
 \affil[2]{Pabna University of Science and Technology, Pabna, 6600, Bangladesh}
 \affil[3]{\orgdiv{Department of CSE}, \orgname{Rajshahi University}, \orgaddress{, \city{Rajshahi}}}
\affil[4]{\orgdiv{School of Computer Science and Engineering, University of Aizu}, \orgaddress{, \city{Aizuwakamatsu, Japan}}}


\abstract{In recent years, advanced artificial intelligence technologies, such as ChatGPT, have significantly impacted various fields, including education and research. Developed by OpenAI, ChatGPT is a powerful language model that presents numerous opportunities for students and educators. It offers personalized feedback, enhances accessibility, enables interactive conversations, assists with lesson preparation and evaluation, and introduces new methods for teaching complex subjects. However, ChatGPT also poses challenges to traditional education and research systems. These challenges include the risk of cheating on online exams, the generation of human-like text that may compromise academic integrity, a potential decline in critical thinking skills, and difficulties in assessing the reliability of information generated by AI.
This study examines both the opportunities and challenges ChatGPT brings to education from the perspectives of students and educators. Specifically, it explores the role of ChatGPT in helping students develop their subjective skills. To demonstrate its effectiveness, we conducted several subjective experiments using ChatGPT, such as generating solutions from subjective problem descriptions. Additionally, surveys were conducted with students and teachers to gather insights into how ChatGPT supports subjective learning and teaching. The results and analysis of these surveys are presented to highlight the impact of ChatGPT in this context.}


\maketitle

\section{Introduction}\label{sec1}
Artificial intelligence has rapidly transformed various sectors over the past few decades, with education being one of its most significant beneficiaries. OpenAI's ChatGPT, a conversational AI model, has emerged as a key tool with the potential to revolutionize education, particularly in developing countries \cite{openai}. By utilizing natural language processing, ChatGPT generates human-like responses, offering innovative ways to provide educational support and create guidelines.
According to the World Bank (2024) \cite{worldbankOverview}, developing countries face significant challenges, including limited resources and inadequate infrastructure. The lack of access to quality education and trained teachers are major contributors to these challenges. ChatGPT can be crucial in bridging the educational gap by expanding learning opportunities and improving access to quality educational materials in underserved areas.
In developing countries, ChatGPT presents significant potential for enhancing education for both learners and teachers. By analyzing large volumes of academic data, ChatGPT identifies trends and gaps that can help policymakers make informed, evidence-based decisions \cite{benboujja2024overcoming}. 

This technology addresses challenges like instant feedback, personalized learning, and academic support, making education more accessible, especially for those with geographic or economic barriers \cite{10184267}. Unlike traditional classrooms, where one teacher handles many students, ChatGPT offers tailored content that adapts to each learner's pace and style \cite{dimitriadou2023critical}. For teachers, it automates tasks, generates materials, and provides additional tutoring support, allowing them to focus on individualized instruction \cite{dimitriadou2023critical}. ChatGPT’s reach also extends to educational policy and guideline development, while its multilingual features enable learning in preferred languages, which is crucial for countries with diverse linguistic backgrounds. However, ethical and practical concerns must be addressed for effective implementation \cite{kuleto2021exploring}. To maximize ChatGPT's impact on education, developing countries need to address data privacy, digital access, and teacher training. Privacy measures are essential to protect student data, and affordable internet access and devices are crucial for equitable participation \cite{unescoBridgingEducational}. Public-private partnerships can help expand digital infrastructure, and teachers require training to use AI tools like ChatGPT effectively \cite{holmes2019artificial}. Addressing these challenges can enhance education across all levels, supporting both educational and economic growth \cite{openai}).
Over the past year, many researchers have analyzed ChatGPT usage among students, teachers, and educators. Rahman et al. conducted a detailed study on ChatGPT's role in education, highlighting its use in personalized feedback, accessibility, interactive learning, lesson preparation, and teaching complex concepts, particularly in programming \cite{rahman2023chatgpt}. They also identified risks, such as cheating, reduced critical thinking, and challenges in evaluating AI-generated content. The study tested ChatGPT's programming assistance through code generation, pseudocode, and code correction, which were validated with an online judge system. Additionally, surveys gathered insights from students and teachers on ChatGPT's effectiveness.
Similarly, Lo et al. explored ChatGPT’s early challenges within the first three months of release \cite{lo2023impact}. Recent research also examines how students use ChatGPT to support learning \cite{Nisar2023, Pavlik2023Collaborating, rudolph2023chatgpt, Fijacko2023, Hargreaves2023, cotton2023chatting}. Studies focusing on educators reveal ChatGPT’s potential in supporting teaching routines and material preparation \cite{topsakal2022framework, zhai2023chatgpt, baidoo2023education, Wang2023, worldbankOverview}.

While ChatGPT offers numerous benefits for learners and educators, including enhanced learning experiences and streamlined content creation, it also presents challenges. A key concern is the risk of misinformation and bias in AI-generated content. Recent studies highlight issues like potential plagiarism and the spread of inaccuracies \cite{mbakwe2023chatgpt, baidoo2023education, khalil2023, Susnjak2022ChatGPT, Choi2023, Hargreaves2023, szabo2023chatgpt, perkins2023academic, rudolph2023chatgpt}. Although these studies provide insight into ChatGPT's limitations, they often lack detailed analysis tailored to various educational contexts.
Research has also examined ChatGPT's use in specific subject areas, such as Economics \cite{Geerling2023}, English Language \cite{deWinter2023}, Law \cite{Hargreaves2023}, Sports Science \cite{szabo2023chatgpt}, Medical Education \cite{Kung2023}, Higher-Order Thinking \cite{Susnjak2022ChatGPT}, Mathematics \cite{Frieder2023}, Programming \cite{Buchberger2023, rahman2023chatgpt}, and Software Testing \cite{jalil2023chatgpt}. However, there is a gap in understanding ChatGPT's role in programming and engineering-related subjects. Little research has focused on how engineering students specifically use ChatGPT, their motivations, or the reliability of the AI content they rely on. Addressing this gap is crucial to better understand and support the unique needs of engineering students.
To address the challenges and gaps identified in prior research, our study provides an in-depth analysis of ChatGPT's use among engineering students and educators, examining how they can use the tool, their motivations, and their views on its reliability. This focused investigation sheds light on ChatGPT's unique role in engineering education, enriching our understanding of its impact in this field. Additionally, we emphasize the strategic integration of ChatGPT in education, particularly in emerging and developing countries, by analyzing practical applications and their implications. Our key contributions are as follows:
\begin{itemize}
\item 
First, we provide a thorough analysis of how ChatGPT is being deployed in these education systems, highlighting its potential benefits, such as enhancing access to and quality of education, while also addressing the associated risks and ethical considerations. Our work goes beyond the scope of existing studies by offering concrete solutions to the challenges they identified.
\item We collected and analyzed data from engineering students and educators to understand their specific use cases, motivations for using ChatGPT, and their perceptions of its reliability. This approach helps identify usage patterns and concerns unique to the engineering domain. Through real-world analysis, we provide actionable insights and practical recommendations for learners and educators, addressing ChatGPT-specific challenges in this field.
Our dataset included responses from engineering students and teachers, with 15 survey questions covering ChatGPT usage in academic tasks such as research, problem-solving, programming, and essay writing. The survey aimed to evaluate the perceived benefits, challenges, and overall effectiveness of ChatGPT in education. Statistical analysis of the responses offers insights that can guide ChatGPT users and inform best practices for its application across various educational sectors.
\item 
Our findings aim to guide policymakers, educators, and stakeholders in effectively integrating ChatGPT into their educational frameworks, maximizing its benefits while minimizing potential drawbacks. By shedding light on these critical aspects, our study not only supports educational advancement but also fosters a more informed and strategic approach to leveraging AI tools in the engineering education sector.
\end{itemize}

\section{Literature Review}
ChatGPT attracts students, educators, researchers, and the general public for its in-depth knowledge across diverse subjects. Despite its benefits, concerns remain about copyright and potential misuse. To address these, researchers examine user engagement and ChatGPT's content generation for educational support and subject-specific applications. Many researchers have been conducting comprehensive analyses of ChatGPT, focusing on its use for learning and teaching, providing subject-specific solutions, and addressing concerns related to copyright and plagiarism issues. ChatGPT can act as a virtual tutor, supporting students' learning in a variety of ways.
Researchers have analyzed the impact of ChatGPT on student learning, as seen in Table \ref{tab:student_learning_teacher_educators}, which categorizes ChatGPT's functions into six main areas: Question Answering, Information Summarization, Exam Preparation, Draft Assistance, and Providing Feedback.  Rudolph et al. \cite{rudolph2023chatgpt} highlight how ChatGPT can structure discussions and guide group interactions, making debates more productive \cite{kasneci2023chatgpt}. Gilson et al. \cite{Gilson2023} found that this improves problem-solving and learning outcomes. Rahman et al. \cite{rahman2023chatgpt} analyze how ChatGPT aids learners in developing programming and problem-solving skills. In assessments, students use ChatGPT to refine drafts and improve content quality \cite{Gilson2023}. Its responses can encourage students to ask deeper questions, promoting critical thinking and knowledge application. However, as noted by Rudolph et al. \cite{rudolph2023chatgpt}, while ChatGPT is a helpful learning aid, it should complement—not replace—students' critical thinking and original work.

\begin{table}[!htp]
\centering
\caption{ChatGPT functions to support student learning and educators \cite{lo2023impact}.} \label{tab:student_learning_teacher_educators}
\setlength{\tabcolsep}{2pt}
\begin{tabular}{|p{2cm}|l|l|p{8cm}|p{1.5cm}|}
\hline
\textbf{Authors}   & \textbf{Year} &   \begin{tabular}[c]{@{}l@{}} \textbf{Student Learning}\\\textbf{Function}\end{tabular}                                      & \textbf{Their Comments }                                                                                               & \textbf{Other Studies} \\ \hline
Nisar et al. \cite{Nisar2023}        & 2023          & Question and Answering                          & ChatGPT provided relevant, accurate answers, making it a useful tool for quick reference and self-study. \cite{Nisar2023} (p. 1).        & \cite{Gilson2023,baidoo2023education}   \\ \hline
Pavlik et al. \cite{Pavlik2023Collaborating}          & 2023          & \begin{tabular}[c]{@{}l@{}}Summarising \\information\end{tabular}                       & ChatGPT excels at processing, distilling, and verbally presenting information. \cite{Pavlik2023Collaborating} (p. 92).                                                             & \cite{kasneci2023chatgpt,Wang2023,Hargreaves2023}                       \\ \hline
Rudolph et al. \cite{rudolph2023chatgpt}          & 2023          & \begin{tabular}[c]{@{}l@{}}Facilitating \\collaboration\end{tabular}                     & `ChatGPT can generate scenarios that enable students to collaborate on problem-solving and goal achievement' \cite{rudolph2023chatgpt} (p. 13). & \cite{kasneci2023chatgpt,cotton2023chatting,Gilson2023}                        \\ \hline
Fijacko et al. \cite{Fijacko2023}          & 2023          & \begin{tabular}[c]{@{}l@{}}Concept checking \\and exam preparation\end{tabular}      & ChatGPT has demonstrated potential as a strong reference and self-learning tool for preparing life support exams \cite{Fijacko2023} (p. 1).                            & \cite{Nisar2023,Mogali2023,Choi2023}                         \\ \hline
   \begin{tabular}[c]{@{}l@{}}Hargreaves \\et al. \cite{Hargreaves2023}\end{tabular}        & 2023          & \begin{tabular}[c]{@{}l@{}}Drafting \\assistance\end{tabular}                            & Students could be encouraged to use AI to generate a 'first draft' response, which they can then refine and improve manually. \cite{Hargreaves2023} (p. 21).                              & \cite{Choi2023,Gilson2023,zhang2023preparing}                        \\ \hline
Cotton et al. \cite{cotton2023chatting}          & 2023          & \begin{tabular}[c]{@{}l@{}}Providing \\feedback\end{tabular}                             & ChatGPT can be utilized to grade assignments and offer real-time feedback to students.\cite{cotton2023chatting} (p. 2).                                                      &\cite{baidoo2023education,atlas2023chatgpt,perkins2023academic}                        \\ \hline

  &  &   \begin{tabular}[c]{@{}l@{}} \textbf{Teacher Related}\\\textbf{Function}\end{tabular}                                      & \                                                                                               &  \\ \hline
  
Topsakal et al. \cite{topsakal2022framework}          & 2023          & \begin{tabular}[c]{@{}l@{}}Generating \\course materials\end{tabular}                  & They queries to generate one of the dialogues in a format compatible with DialogFlow, and it successfully completed the task \cite{topsakal2022framework} (p. 37). &\cite{ali2023devilspeak,kasneci2023chatgpt,khan2023chatgpt}                             \\ \hline
Zhai et al. \cite{zhai2023chatgpt}          & 2023          & \begin{tabular}[c]{@{}l@{}}Providing \\suggestions\end{tabular}                       & They asked it that the learner had dyslexia, and ChatGPT eventually suggested specific learning materials tailored for the learner \cite{zhai2023chatgpt} (p. 1).                          & \cite{megahed2023generative,Han2023}                         \\ \hline
    \begin{tabular}[c]{@{}l@{}}Baidoo-Anu \\et al. \cite{baidoo2023education}  \end{tabular}       & 2023          & \begin{tabular}[c]{@{}l@{}}Performing language \\translation\end{tabular}              & ChatGPT can assist in translating educational materials into various languages. \cite{baidoo2023education} (p. 8).                                                                           & \cite{khan2023chatgpt,bishop2023computer,atlas2023chatgpt}                        \\ \hline
Wang et al. \cite{Wang2023}          & 2023          & \begin{tabular}[c]{@{}l@{}}Generating \\assessment tasks\end{tabular}                  & ChatGPT can also generate exercises, quizzes, and scenarios to support classroom practice and assessment. \cite{Wang2023} (p. 5).   &\cite{Han2023,khan2023chatgpt,alworafi2023ai}                         \\ \hline
Wang et al. \cite{Wang2023}          & 2023          & \begin{tabular}[c]{@{}l@{}}Evaluating student \\performance\end{tabular}               & ChatGPT can be trained to grade student essays, allowing teachers more time to focus on other aspects of instruction \cite{Wang2023} (p. 8).                                    & \cite{baidoo2023education,cotton2023chatting,qadir2023engineering}                        \\ \hline
\end{tabular}
\end{table}
\begin{table}[!htp]
\centering
\caption{Major potential issues including threat and plagiarism associated with ChatGPT \cite{lo2023impact}.} \label{tab:threat_plagiarism}

\setlength{\tabcolsep}{3pt}
\begin{tabular}{|p{2cm}|l|l|p{8cm}|p{1.5cm}|}
\hline
\textbf{Authors} & \textbf{Year} & \textbf{Issues} & \textbf{Representative Quotes} & \textbf{Other Studies} \\ \hline
Mbakwe et al. \cite{mbakwe2023chatgpt} & 2023 & \begin{tabular}[c]{@{}l@{}}Relying on \\ biased data\end{tabular} & These biases arise from research conducted in high-income countries and the textbooks used to train ChatGPT. \cite{mbakwe2023chatgpt} (p. 2). & \cite{baidoo2023education,tlili2023chatgpt,Pavlik2023Collaborating} \\ \hline
Baidoo-Anu et al. \cite{baidoo2023education} & 2023 & \begin{tabular}[c]{@{}l@{}}Having limited \\ up-to-date knowledge\end{tabular} & ChatGPT lacks knowledge of events after 2021, so it cannot provide references or information beyond that year. \cite{baidoo2023education} (p. 14). & \cite{khalil2023,perkins2023academic,Gilson2023} \\ \hline
Baidoo-Anu et al. \cite{baidoo2023education} & 2023 & \begin{tabular}[c]{@{}l@{}}Generating incorrect/\\fake information\end{tabular} & ChatGPT generated a fictitious article, complete with fabricated bibliographic details and a non-functional URL. \cite{baidoo2023education} (p. 14). & \cite{Mogali2023,jalil2023chatgpt,king2023conversation} \\ \hline
Baidoo-Anu et al. \cite{baidoo2023education} & 2023 & \begin{tabular}[c]{@{}l@{}}Student \\ plagiarism\end{tabular} & Their team used ChatGPT, encountered slightly more plagiarism issues compared to the control group that did not use ChatGPT.' (p. 7). & \cite{tlili2023chatgpt,qadir2023engineering,ventayen2023} \\ \hline
Khalil et al. \cite{khalil2023} & 2023 & \begin{tabular}[c]{@{}l@{}}Bypassing plagiarism \\ detectors\end{tabular} & Out of the 50 essays reviewed, the plagiarism found 40  and that is 20\%. \cite{khalil2023} (p. 10). & \cite{ventayen2023,rudolph2023chatgpt,stutz2023chatgpt} \\ \hline
Susnjak et al. \cite{Susnjak2022ChatGPT} & 2023 & \begin{tabular}[c]{@{}l@{}}Incorporating \\ multimedia resources\end{tabular} & Embedding images to exam questions can make it more difficult for students to cheat and for ChatGPT to generate accurate responses \cite{Susnjak2022ChatGPT} (p. 16). & \cite{Geerling2023,Newton2023,rudolph2023chatgpt} \\ \hline
Choi et al. \cite{Choi2023} & 2023 & \begin{tabular}[c]{@{}l@{}}Adopting novel \\ question types\end{tabular} & Instructors should rethink the types of questions they ask and based on the legal rules.\cite{Choi2023} (p. 12). & \cite{Geerling2023} \\ \hline
Hargreaves et al. \cite{Hargreaves2023} & 2023 & \begin{tabular}[c]{@{}l@{}}Employing digital-free \\ assessment formats\end{tabular} &  Teacher can make Blanket solution in all assessments of the ‘in-class’ variety and that not possible in ChatGPT for some condition \cite{Hargreaves2023} (p. 19). & \cite{Susnjak2022ChatGPT,Hisan2023,stutz2023chatgpt} \\ \hline
Szabo et al. \cite{szabo2023chatgpt} & 2023 & \begin{tabular}[c]{@{}l@{}}Using AI-based \\ writing detection tools\end{tabular} & Traditional plagiarism detectors failed to recognize AI-generated text, AI-specific detectors were able to identify it \cite{szabo2023chatgpt} (p. 2). & \cite{Susnjak2022ChatGPT,zhang2023preparing,perkins2023academic} \\ \hline
Perkins et al. \cite{perkins2023academic} & 2023 & \begin{tabular}[c]{@{}l@{}}Checking \\ references\end{tabular} & Although in-text citations and references were included, they were entirely fabricated, which provides a potential way for academic staff to detect the use of AI-generated content. \cite{perkins2023academic} (p. 5). & \cite{rudolph2023chatgpt,cotton2023chatting,szabo2023chatgpt} \\ \hline
Choi et al. \cite{Choi2023} & 2023 & \begin{tabular}[c]{@{}l@{}}Establishing anti-\\ plagiarism guidelines\end{tabular} & Administrations should rethink honour codes to address and regulate the use of language models. \cite{Choi2023} (p. 12). & \cite{rudolph2023chatgpt,perkins2023academic,khalil2023} \\ \hline
Rudolph et al. \cite{rudolph2023chatgpt} & 2023 & \begin{tabular}[c]{@{}l@{}}Providing student \\ education\end{tabular} & Authors recommend students stay informed about academic integrity policies, understand the consequences of academic misconduct, and receive proper training on academic integrity. \cite{rudolph2023chatgpt} (pp. 14–15). & \cite{garcia2023perception,willems2023chatgpt,Pavlik2023Collaborating}\\ \hline
\end{tabular}
\end{table}

\begin{table*}[!htp]
\centering
\caption{ ChatGPT-3.5 performance evaluation across various subject domains \cite{lo2023impact}}\label{tab:subjective_domain_oriented}
\begin{adjustwidth}{-2cm}{0cm}
\setlength{\tabcolsep}{2pt}

\begin{tabular}{|p{2cm}|p{1cm}|p{2.5cm}|p{2.5cm}|p{7.5cm}|p{1.5cm}|}
\hline
\textbf{Authors} & \textbf{Year} & \textbf{Subject Domain} & \textbf{Overall Performance} & \textbf{Researcher Comments} & \textbf{Other Studies}  \\ \hline
Geerling et al. \cite{Geerling2023} & 2023 & Economics & Outstanding & ChatGPT ranked in the 99th percentile for macroeconomics and 91st percentile for microeconomics compared to students. & - \\ \hline
De Winter et al. \cite{deWinter2023} & 2023 & English language comprehension & Satisfactory & ChatGPT's mean grade of 7.18 was similar to the average student performance in the Netherlands. & - \\ \hline
Hargreaves et al. \cite{Hargreaves2023} & 2023 & Law & Barely satisfactory to unsatisfactory & ChatGPT performed at the level of a C+ student and struggled most with problem-style or ‘issue spotting’ questions. & \cite{Choi2023} \\ \hline
Szabo et al. \cite{szabo2023chatgpt} & 2023 & Sports science and psychology & Unsatisfactory & ChatGPT answered some questions correctly but failed the test with a 45\% score. & - \\ \hline
Kung et al. \cite{Kung2023} & 2023 & Medical Education & Barely satisfactory to unsatisfactory & ChatGPT showed moderate accuracy in USMLE but failed American Heart Association exams. & \cite{Kung2023,Gilson2023,Fijacko2023,Han2023} \\ \hline
Susnjak et al. \cite{Susnjak2022ChatGPT} & 2023 & Critical and higher-order thinking & Outstanding & Responses were clear, precise, and relevant to requests. & - \\ \hline
Frieder et al. \cite{Frieder2023} & 2023 & Mathematics & Unsatisfactory & ChatGPT's math abilities were significantly below those of an average graduate student. & - \\ \hline
Buchberger et al. \cite{Buchberger2023} & 2023 & Programming & Outstanding to satisfactory & Most answers were correct and well-explained, but an assignment was graded only ‘Satisfactory.’ & \cite{stutz2023chatgpt,megahed2023generative} \\ \hline
Jalil et al. \cite{jalil2023chatgpt} & 2023 & Software testing & Unsatisfactory & ChatGPT answered 37.5\% of questions correctly, insufficient to pass a software testing course. & - \\ \hline
Newton et al. \cite{Newton2023} & 2023 & MCQ-based exams across subjects & Unsatisfactory & ChatGPT failed most MCQ exams and performed worse than the average human student. & - \\ \hline
\end{tabular}
\end{adjustwidth}
\end{table*}

ChatGPT provides valuable support for teachers and instructors in both the preparation and assessment phases as shown in Table \ref{tab:student_learning_teacher_educators}. Its main applications are in teaching preparation—including generating course materials, offering suggestions, and translating content—and assessment—creating tasks and evaluating student performance. 
Concerns have been raised about ChatGPT's ability to produce polished but inaccurate information, as shown in Table \ref{tab:threat_plagiarism}. Mogali \cite{Mogali2023} and others \cite{szabo2023chatgpt,baidoo2023education,perkins2023academic,khalil2023} highlight that ChatGPT often generates incorrect content, including fake citations, which is particularly problematic in academia, where accuracy is crucial. Megahed et al. \cite{megahed2023generative} found that ChatGPT can produce flawed code without recognizing errors, a concern echoed by Jalil et al. \cite{jalil2023chatgpt}, who noted its limited ability to judge its own accuracy. This issue extends across fields such as mathematics \cite{Frieder2023}, sports science \cite{szabo2023chatgpt}, and health professions \cite{cotton2023chatting,bishop2023computer,Mogali2023}, raising concerns about its reliability. Another issue is ChatGPT’s potential to bypass plagiarism detection. Ventayen \cite{ventayen2023} found that ChatGPT-generated essays yielded a low similarity score on Turnitin, indicating minimal detectable plagiarism. Khalil and Er \cite{khalil2023} observed similar results, with an average similarity score of 13.72\% on Turnitin and 8.76\% on iThenticate \cite{ithenticate2023}, suggesting ChatGPT’s text often appears original and may challenge academic integrity. To mitigate misuse, researchers propose alternative assessment methods. Zhai \cite{zhai2023chatgpt} recommends creative assignments that encourage critical thinking, while Choi et al. \cite{Choi2023} suggests focusing on case analysis over rote knowledge recall. Geerling et al. \cite{Geerling2023} propose tasks that require students to produce AI-resistant materials, and Stutz et al. \cite{stutz2023chatgpt} emphasize higher-order skills in line with Bloom’s taxonomy \cite{bloom1956}. AI-specific plagiarism detectors also show promise in flagging AI-generated content \cite{szabo2023chatgpt}, and ChatGPT’s often inaccurate reference lists \cite{cotton2023chatting,qadir2023engineering} can aid in identifying potential misuse. To address these issues, researchers stress the need for clear anti-plagiarism guidelines and educating students on academic integrity \cite{rudolph2023chatgpt}.
In addition to exploring student learning and teaching applications, it is essential to analyze ChatGPT's use in specific academic fields. Researchers have assessed ChatGPT's performance across various disciplines, including law, mathematics, and medical education. Table \ref{tab:subjective_domain_oriented} lists authors and their findings. Most studies, except de Winter's \cite{deWinter2023} high school exam analysis, focused on higher education. Results showed ChatGPT performed well in critical thinking and economics \cite{Susnjak2022ChatGPT,Geerling2023} but struggled in law \cite{Choi2023, Hargreaves2023}, medical education \cite{Kung2023, Gilson2023, Fijacko2023, Han2023}, and mathematics \cite{Frieder2023}. Newton’s study \cite{Newton2023} revealed that ChatGPT excelled in economics but scored 8–40 points lower than average students in other fields. In medical education, Kung et al. \cite{Kung2023} and Gilson et al. \cite{Gilson2023} found that ChatGPT passed the US Medical Licensing Examination (USMLE) with moderate accuracy, but Fijačko \cite{Fijacko2023} noted it failed the American Heart Association's life support exams. Han et al. \cite{Han2023} also reported incomplete information from ChatGPT on cardiovascular diseases. In Malaysia, Nisar and Aslam \cite{Nisar2023} observed that ChatGPT provided accurate pharmacology answers but lacked proper references. Similarly, ChatGPT scored below average on medical exams in China \cite{Wang2023}, Korea \cite{Huh2023}, India \cite{Hisan2023}, Singapore \cite{Mogali2023}, and Bangladesh \cite{rahman2023chatgpt}. Overall, these findings suggest that while ChatGPT shows promise in certain areas, its performance in medical education and other specialized fields remains limited.

\section{ChatGPT in Research and Education: Our Real-Time Command for Exploring Benefits and Threats}
\begin{figure}
  \begin{adjustwidth}{-3cm}{0cm}
    \centering
    \includegraphics[width=0.8\linewidth]{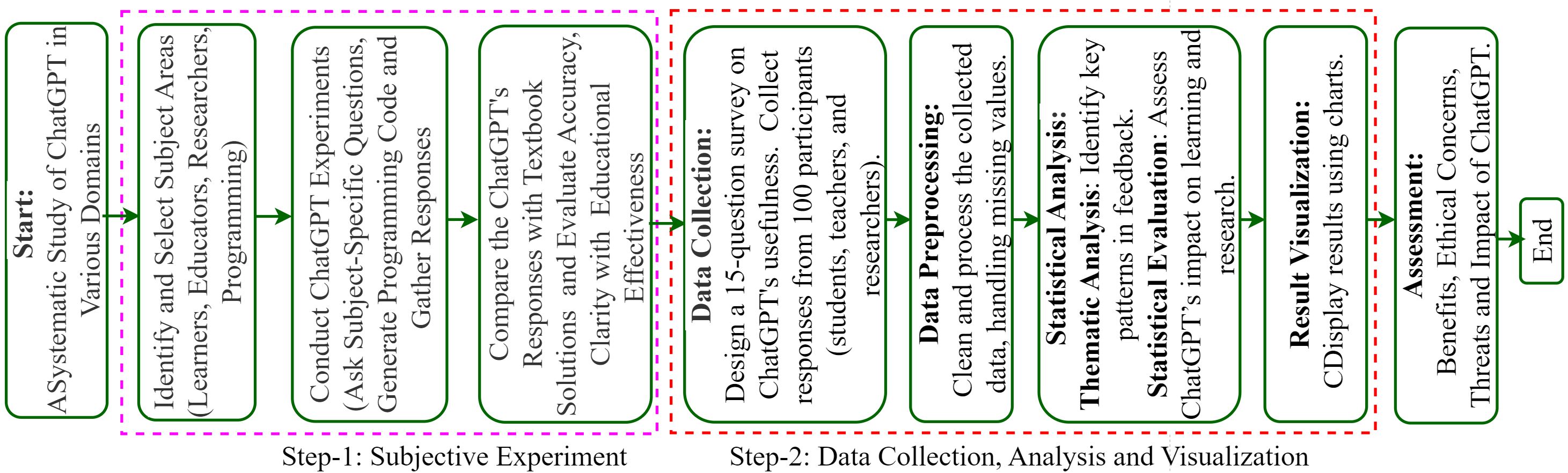}
    \caption{Abstract view of proposed methodology of ChatGPT in Research and Education}
    \label{fig:abstractviewofpaper}
    \end{adjustwidth}
\end{figure}

The study systematically examines the benefits and risks of ChatGPT in research and education. It focuses on four areas: opportunities and challenges for learners, educators, and researchers, and its use in programming education. The approach includes experiments and surveys to collect data from students and teachers.
An abstract representation of the proposed methodology is illustrated in Figure ~\ref{fig:abstractviewofpaper}.

\begin{itemize} [left=0cm]
    \item \textbf{Opportunities for Learners}: 
       This section assesses ChatGPT’s ability to solve subject-specific problems and compares its answers to established solutions found in textbooks. This comparison highlights ChatGPT’s effectiveness as a learning aid.
    
    \item \textbf{Opportunities for Educators}: This section assesses ChatGPT’s capacity to assist in lesson planning, answering scientific questions, and providing explanations of complex topics like Newton's laws and chemistry.
      
    \item \textbf{Opportunities for Researchers}: This section explore ChatGPT’s potential to aid in academic writing, idea generation, literature review, and data analysis by showcasing examples of how it can be used in research workflows.     
    \item \textbf{Programming Learning with ChatGPT}: This section evaluate ChatGPT's ability to explain programming concepts and provide working code, assessing its role as a tool for learning programming.
\end{itemize}
We categorized ChatGPT commands and queries based on user types: students, teachers, researchers, and programmers, summarizing the findings for each group. To evaluate ChatGPT's role in education and research, we developed an experimental framework to assess its effectiveness in solving subject-specific problems, its reliability in providing educational support, and its perceived value to authors.

\vspace{3mm} 

Through these analyses, we aim to provide insights into the integration of ChatGPT in educational frameworks.

\begin{figure}
    \centering
    \includegraphics[width=.60\linewidth]{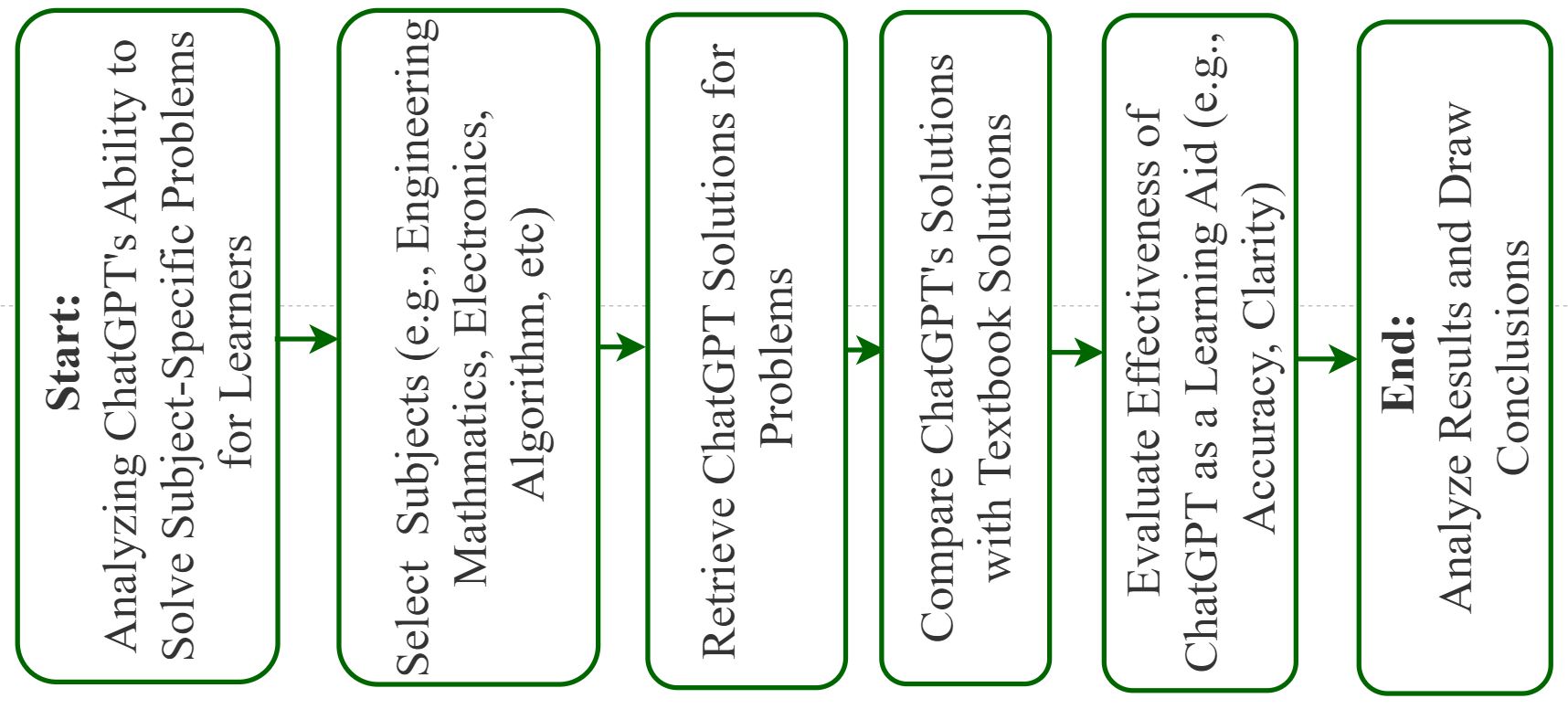}
    \caption{Opportunities and challenges for Learners with ChatGPT}
    \label{fig:ChatGPT_Learners}
\end{figure}

\begin{figure}[h]
\centering
\includegraphics[width=0.6\textwidth]{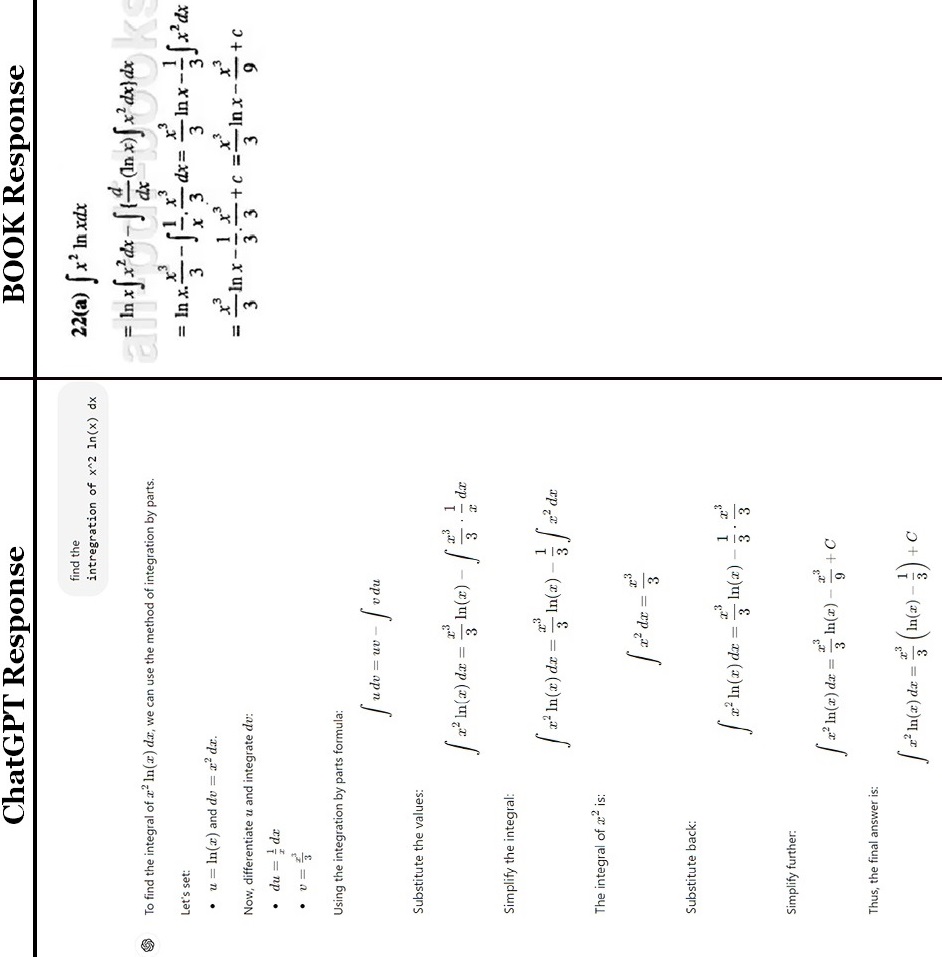}
\caption{ChatGPT's Capabilities in Technical Education, Solving mathematical problems.}\label{fig:integrate_equation}
\end{figure}


\subsection{Opportunities for Learners}
In this subsection, we visualized how ChatGPT enhances engineering students' learning by simplifying complex concepts in mathematics, programming, and computer science. It offers personalized assistance, supports skill development, facilitates group discussions, and improves accessibility. By comparing ChatGPT's responses with textbook solutions, the analysis highlights its potential as a valuable tool for mastering advanced engineering topics. The detailed procedure is illustrated in Figure ~\ref{fig:ChatGPT_Learners} Step-1.

\subsubsection{Enhanced Learning Experience and Skill Development With Dynanic Solution}
ChatGPT enhances engineering students' learning experience, especially in mathematics, programming, and computer science. It simplifies complex concepts with step-by-step explanations and real-world analogies. For instance, a student can ask for both an explanation and a code example for sorting algorithms or get a breakdown of mathematical concepts like integration. This personalized support helps learners tackle difficult topics effectively, regardless of their expertise level.
For instance, a student studying QuickSort might initially struggle with understanding the partitioning process. By querying ChatGPT, they could receive an explanation like: "QuickSort works by selecting a 'pivot' element and partitioning the other elements into two sub-arrays according to whether they are less than or greater than the pivot. These sub-arrays are then recursively sorted." Additionally, ChatGPT can provide Python code for QuickSort, helping the student not only visualize but also implement the algorithm.
To assess ChatGPT’s practical utility for engineering students, we compared its answers to well-documented textbook solutions in two specific domains: mathematics and physics.

\begin{figure}[h]
\centering
\includegraphics[width=1.05\textwidth]{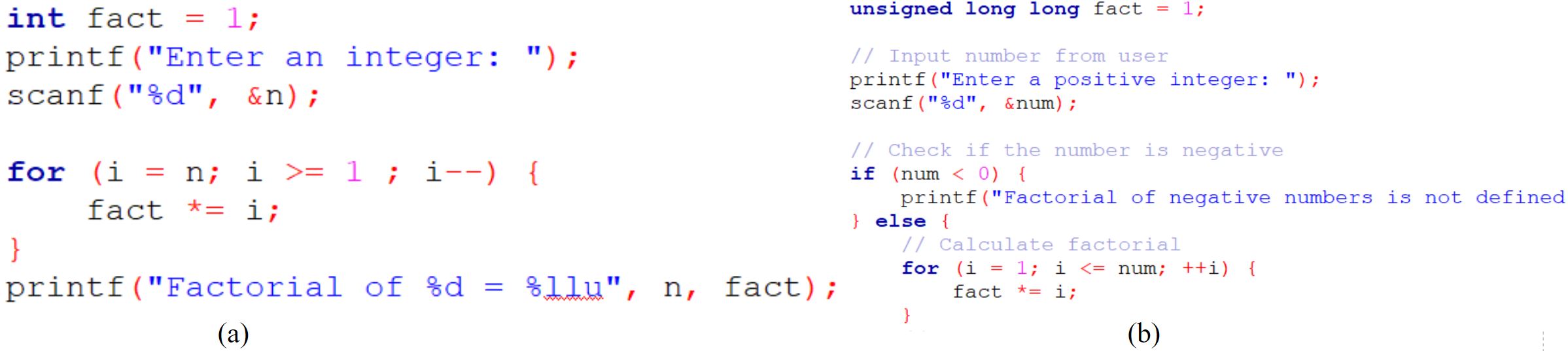}
\caption{Calculate the factorial of a small integer number (a) Freshman's response (b) ChatGPT response.}
\label{fig:factorial_comparison}
\end{figure}

In Figure \ref{fig:integrate_equation}, we asked ChatGPT to solve the integral \(\int x^2 \ln(x)\, dx\), a classic problem covered in mathematical textbooks. By comparing the ChatGPT response to the traditional textbook solution, we can evaluate how closely it aligns with established methods and whether its explanation is suitable for learners, particularly in helping them understand the steps involved in integration. Through these analyses, we aim to showcase the advantages of using ChatGPT as an educational tool in engineering disciplines. Our findings reveal that ChatGPT is not only a resource for basic queries but also a potential guide for mastering challenging engineering concepts, making it a valuable asset for learners. ChatGPT also aids in skill development for engineering students by refining coding skills, offering code suggestions, explaining syntax errors, and suggesting alternative solutions. A student can use ChatGPT to improve code efficiency or troubleshoot bugs. In computer science, it can generate practice exercises and quizzes, helping students build problem-solving and analytical skills based on their specific needs.
For instance, Imagine a student working on a Python project involving a search algorithm. The student wants to implement binary search but is unsure about their code's efficiency. They can consult ChatGPT to review their implementation and suggest improvements.
 In addition,  ChatGPT can significantly enhance the learning experience for a beginner in programming. The code shown in Figure~\ref{fig:factorial_comparison} (a), provided by a Computer Science freshman, is functional and concise, calculating the factorial correctly. However, ChatGPT's response shown in Figure~\ref{fig:factorial_comparison} (b) goes beyond functionality, incorporating several best practices that are crucial for a deeper understanding of programming.

Moreover, ChatGPT enhances the code by adding error handling, checking for negative input, and using an appropriate data type (`unsigned long long`) to handle large factorial values, preventing overflow. It also provides step-by-step comments to clarify the logic, making it easier for learners to follow. These improvements help students write professional, error-resistant, and user-friendly code, fostering better programming habits and deepening their understanding. ChatGPT serves as a valuable tool for learning and improving coding skills.
\subsubsection{Enhance Accessibility of the Disable Person As the Learners}
ChatGPT enhances accessibility in education, particularly for engineering students with disabilities. Features like text-to-speech allow visually impaired students to hear coding exercises, while transcriptions help those with hearing impairments follow spoken instructions. Additionally, students can request simpler explanations of technical content, ensuring it's accessible and easy to understand for all learners.

 Example: A student with visual impairments working on a coding assignment can use ChatGPT’s text-to-speech feature to listen to code examples and explanations, enabling them to complete the task without needing to read the text. Similarly, a student who struggles with technical jargon can ask ChatGPT to simplify complex engineering concepts, making learning more inclusive and personalized.

\subsubsection{Interactive Learning and Group  Discussion}
\begin{figure}
    \centering
    \includegraphics[width=1.25\linewidth]{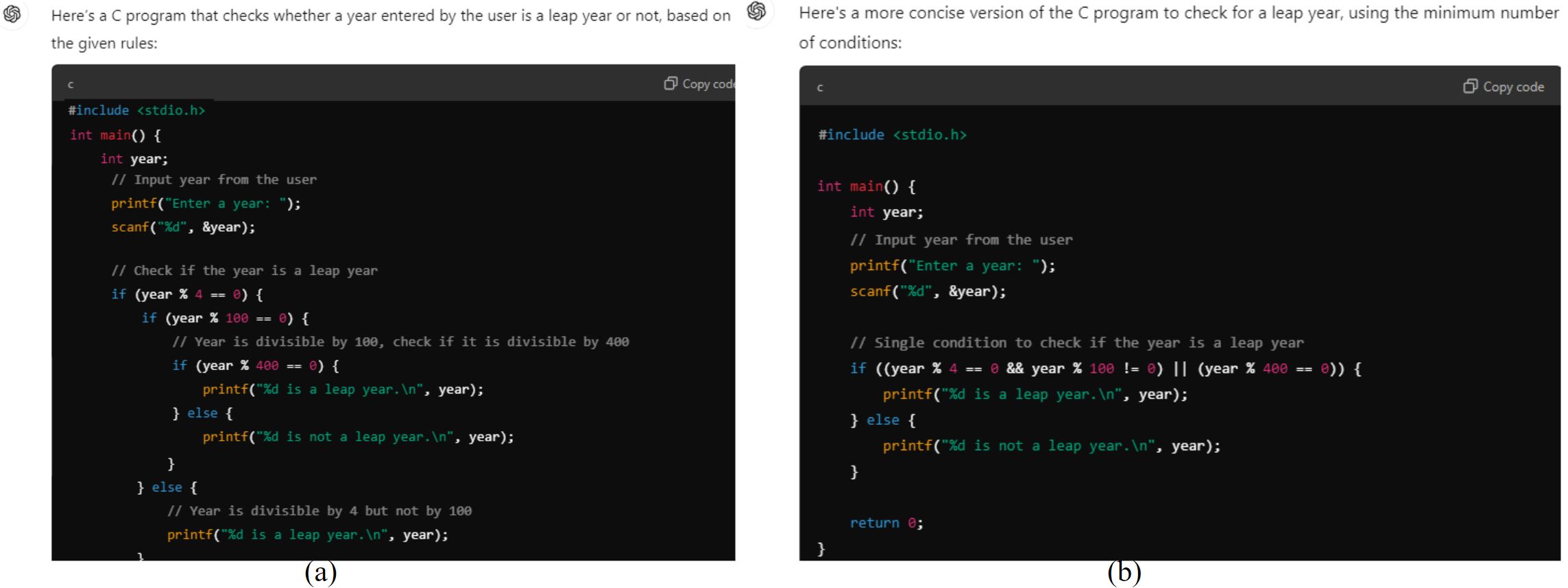}
    \caption{A C Program To Check Leap Year (a)  General Concept (b) Minimal number of conditions }
    \label{fig:leapyear}
\end{figure}

ChatGPT creates an interactive learning environment where engineering students engage in dynamic conversations. For example, a student learning object-oriented programming can ask about differences between classes and objects, followed by questions on inheritance or polymorphism. This interactive approach promotes deeper learning, providing immediate feedback and clarification, which enhances understanding and retention of complex concepts.
\begin{itemize}
    \item \textbf{Example 1:} A student learning about circuit design can ask ChatGPT to explain the differences between series and parallel circuits. After receiving the explanation, they can then follow up with more specific questions about calculating voltage and current in different scenarios, allowing for an active, personalized learning experience that adapts to their understanding in real-time.
    \item \textbf{Example 2:} ChatGPT can significantly enhance the learning experience by enabling interactive learning, which helps learners understand concepts more effectively. Let's explore how ChatGPT aids this process by explaining the differences between the two code examples provided in Figure \ref{fig:leapyear}(a) and \ref{fig:leapyear}(b).
\end{itemize}

In group projects, ChatGPT serves as a collaborative assistant by generating discussion points, providing technical data, and suggesting solutions to engineering problems. For instance, when students debate data structure efficiency, ChatGPT can break down time complexities and suggest optimal structures for specific use cases. It also helps students present arguments clearly and respond to counterpoints, improving both their technical and communication skills.
\begin{itemize}
    \item Example: For instance, during a group discussion on which data structure to use for implementing a priority queue, one student suggests using a binary search tree, while another prefers a heap. To settle the debate, the group consults ChatGPT, which explains the time complexities: O(log n) for both insertion and extraction in a heap versus O(log n) insertion but O(n) extraction for a binary search tree in worst cases. Armed with this information, the group can make a well-informed decision, choosing the heap for optimal performance, and we ask ChatGPT "Write a Python implementation of a priority queue using both a binary search tree and a heap. Compare the performance of insertion and extraction operations between the two implementations for different input sizes. Which implementation performs better for large datasets and why?".
\end{itemize}
The comparison of the two style codes shows in Figure \ref{fig:leapyear} can be explained as below: 
\begin{itemize}[left=2cm]
    \item \textbf{Code Structure and Readability:} In the first code shown in Figure \ref{fig:leapyear}(a), the leap year check is split across multiple \texttt{if-else} statements, examining divisibility by 4, 100, and 400 sequentially. In contrast, the second code shown in Figure \ref{fig:leapyear}(b) condenses this logic into a single line:
    \begin{lstlisting}[language=C]
    (year % 4 == 0 && year % 100 != 0) || (year % 400 == 0)
    \end{lstlisting}
    This concise structure enhances readability.

    \item \textbf{Efficiency:} Both snippets yield the same result, but the second code is more efficient, combining checks into a single \texttt{if} statement and minimizing conditional branches. The first code may perform unnecessary checks if the year is divisible by 4 but not by 100.

    \item \textbf{Simplicity and Maintenance:} The second code’s compact form makes it simpler to understand and maintain, while the first code's nested structure could become harder to manage if expanded.
\end{itemize}

By comparing these snippets, ChatGPT demonstrates how coding style impacts readability and maintainability, helping learners understand efficient code practices.

\subsubsection{Improvement of Assignment and Home Work }
ChatGPT is a valuable tool for students, offering support across various academic subjects. It clarifies complex concepts, provides detailed explanations, and generates ideas for essays or research projects, saving time and enhancing understanding. It also helps structure assignments, improve grammar, and refine arguments for clarity. While it doesn’t replace critical thinking or original research, it guides students through tasks and encourages better learning outcomes. For example, a student can ask ChatGPT to explain control statements or provide examples of loops, making difficult topics more accessible.
\subsection{Opportunities for Educators}
\begin{figure}
    \centering
    \includegraphics[width=0.60\linewidth]{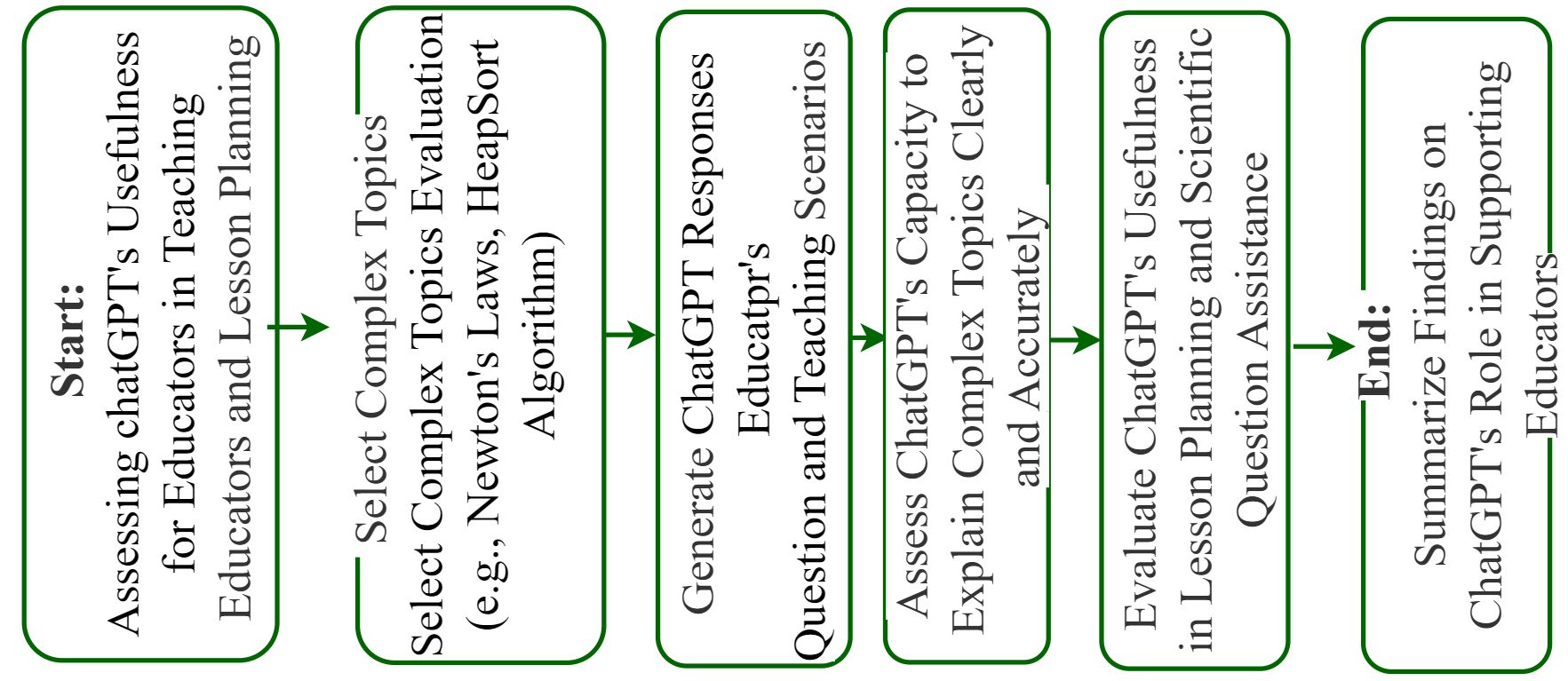}
    \caption{Opportunities and challenges for Educator  with ChatGPT}
    \label{fig:ChatGPT_Educator}
\end{figure}

ChatGPT brings numerous benefits for educators, especially those working with engineering as educators, instructors and teachers. It helps streamline lesson planning, supports personalized learning, offers rapid assessment, and aids in responding to complex student queries. The following sections explore how educators can leverage ChatGPT to enhance teaching and improve student outcomes. "Refer to the detailed process shown in Figure~~\ref{fig:ChatGPT_Educator}.

\subsubsection{Lesson Planning }
\begin{figure}[h]
\centering
\includegraphics[width=1.20\textwidth]{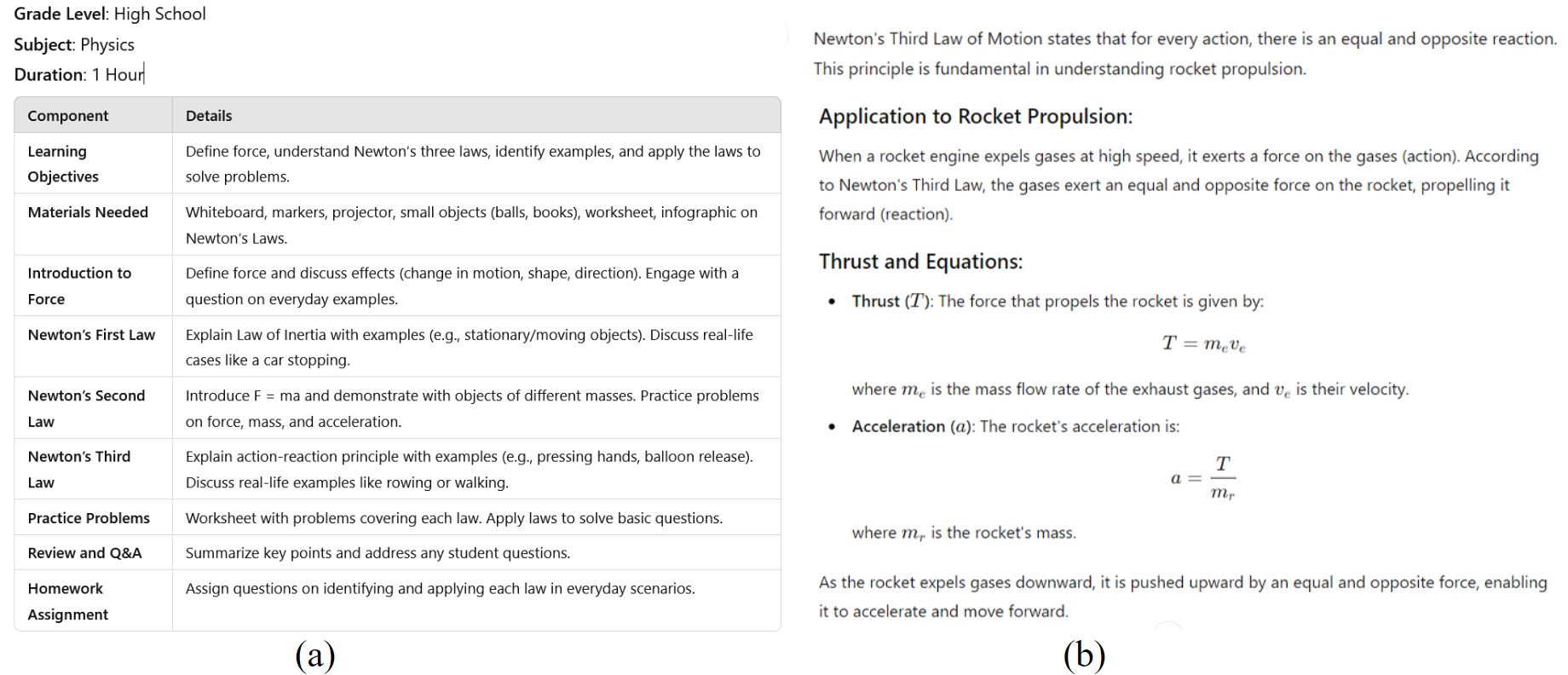}
\caption{ (a)Lesson plan designing using ChatGPT (b) Application ways of scientific law }\label{fig:edu-lesson-plan-design_Sci_applicaiton}
\end{figure}

ChatGPT brings an opportunity to make a comprehensive and efficient lesson design. One of the most time-consuming tasks for educators is developing detailed lesson plans. ChatGPT can greatly assist in generating structured lesson plans that align with curriculum goals.
\begin{itemize}
    \item Example: A physics teacher could ask ChatGPT to "Design a lesson plan for a high school physics class focusing on Newton's law. Please give me a table format and include two columns only components and details" ChatGPT would then provide a structured outline like Figure \ref{fig:edu-lesson-plan-design_Sci_applicaiton} (a). 
\end{itemize}

In the same way, ChatGPT can assist educators across related cross-disciplines, from engineering fields to humanities:
\begin{itemize}
    \item Mathematics: A math instructor could request a lesson plan on calculus, and ChatGPT would break down topics like derivatives, integrals, and limits, offering exercises that cater to both beginner and advanced students.
    \item Language and Literature: An English literature teacher could ask ChatGPT to "Create a lesson plan for teaching Shakespeare's 'Hamlet'," resulting in a comprehensive guide with character analysis, thematic discussions, and historical context.
\end{itemize}
By leveraging ChatGPT for lesson planning, educators can save time while ensuring their lesson plans are thorough, well-organized, and aligned with educational standards.

\subsubsection{Adquate Teaching and Answering Queries}
\begin{figure}[h]
\centering
\includegraphics[width=1.2\textwidth]{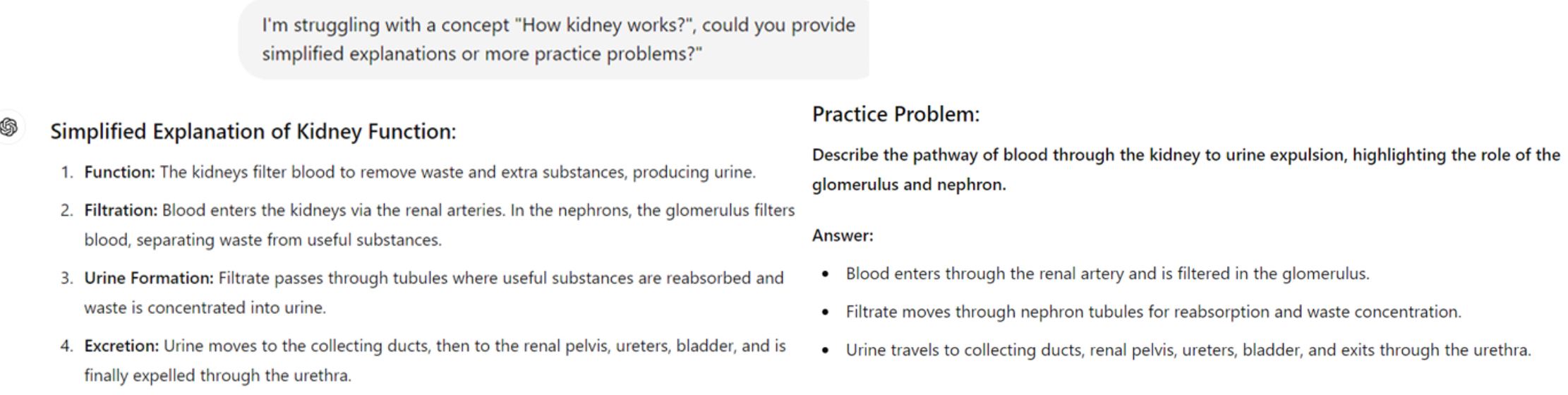}
\caption{Making practice problems with ChatGPT}\label{fig:edu-making-practice-problem-kidney-disease}
\end{figure}

Educators learn at a different paces to do adequate teaching, and personalized support is crucial for ensuring all learners achieve their potential. ChatGPT can be a powerful tool for personalized support by generating custom resources based on a student’s specific needs.
\begin{itemize}
    \item Example: Suppose a student struggles with understanding sorting algorithms in a computer science class. An educator could use ChatGPT to generate a personalized video tutorial or interactive coding exercise that focuses specifically on the types of sorting algorithms the student finds challenging. This targeted assistance helps reinforce the student’s understanding and encourages active learning.
    \end{itemize}
ChatGPT can also be used to adapt learning materials for the teacher in real time based on a student's requirements and progress.
\begin{itemize}
    \item For instance, if a student demonstrates interest in a certain topic and educators lack information about it, ChatGPT can recommend more advanced exercises or additional reading materials to help educators challenge the student further.  Conversely, if an educator struggles, the model can simplify explanations, provide alternative learning methods, or offer more practice problems to build confidence and mastery. Such as interest in any disease or the application of any scientific law.  Figure \ref{fig:edu-making-practice-problem-kidney-disease} show the query of the application of Newton's law and the process of kidney disease where chatGPT generated some crucial information. This information can help the educator explain things to the student clearly. 
This personalized approach enhances educators' student engagement and promotes a deeper understanding of the subject matter.

\end{itemize}

In a classroom setting, students often have questions that require immediate answers. ChatGPT can assist educators by providing accurate, detailed, and contextually relevant responses to student inquiries:
\begin{itemize}
    \item Example: If a student in a physics class asks, "How does Newton's third law apply to rocket propulsion?" ChatGPT can provide a clear explanation that includes the principles of action and reaction forces, along with real-world examples such as the launch of a spacecraft. This enables students to grasp complex concepts more easily and allows educators to address a wider range of questions efficiently. Figure \ref{fig:edu-lesson-plan-design_Sci_applicaiton} (b) shows the screenshot of the question and ChatGPT response. 
\end{itemize}
ChatGPT’s extensive knowledge base makes it particularly useful for answering specialized or complex queries that may require additional research:

\begin{itemize}
    \item For example, in a biology class, a student might ask about the latest research on CRISPR gene-editing technology. ChatGPT can provide an up-to-date summary of current advancements, ethical considerations, and potential applications, helping students stay informed about cutting-edge scientific developments. Figure \ref{fig:edu-making-practice-problem-kidney-disease} shows the query and response from the ChatGPT regarding kidney disease. 
\end{itemize}

By incorporating ChatGPT into the classroom, educators can ensure that student questions are addressed promptly and comprehensively, enhancing the overall learning experience.
\subsubsection{Assessment Material Creation and Rapid Evaluation}
\begin{figure}[h]
\centering
\includegraphics[width=0.8\textwidth]{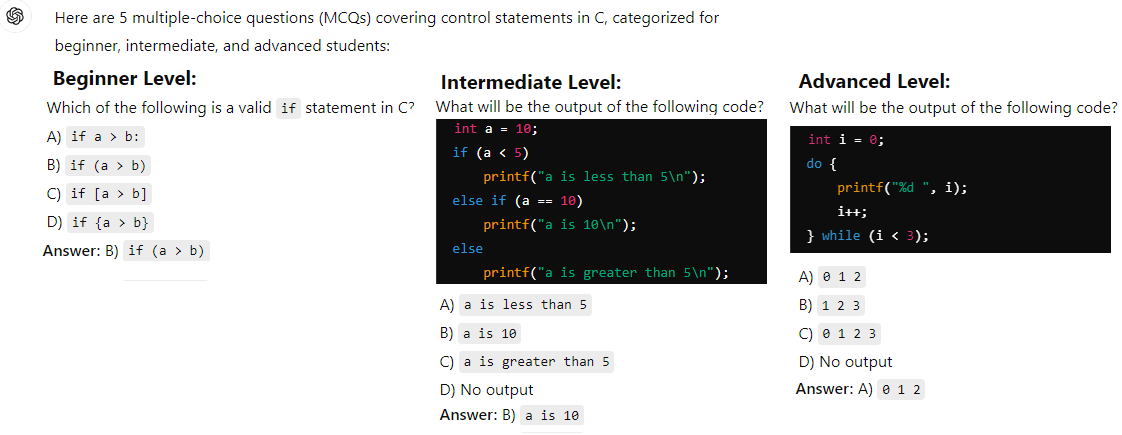}
\caption{ChatGPT Create questions for different
 proficiency levels—beginner, intermediate, and advanced}
\label{fig:educator_programming}
\end{figure}

ChatGPT has the ability to assist educators in preparing assessment materials efficiently by generating a wide range of question types, including multiple-choice questions (MCQs), short answers, and conceptual queries. Specifically, in the example provided:

ChatGPT can create questions for different student proficiency levels—beginner, intermediate, and advanced—based on the topic's complexity (see Figure \ref{fig:educator_programming}).

\begin{figure}[h]
\centering
\includegraphics[width=1.3\textwidth]{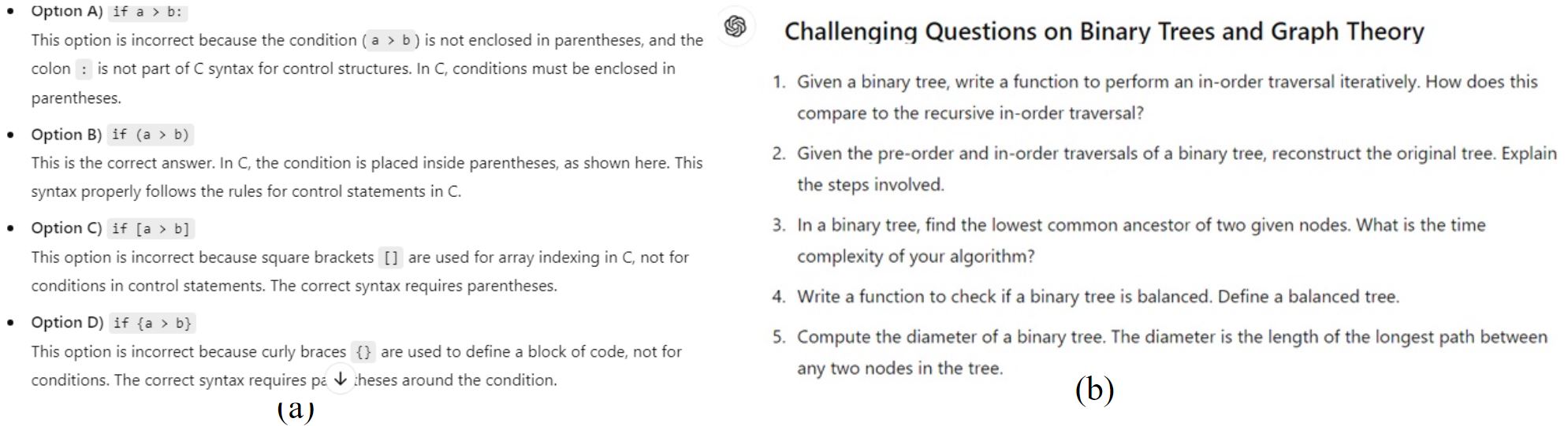}
\caption{ChatGPT's  explanations for (a) The beginning level question (b) Complex question }
\label{fig:educator_question_easy-complex}
\end{figure}

 It ensures diverse coverage of topics, such as control statements in C, while focusing on key concepts (e.g., conditional statements, loops).  ChatGPT provides answers with explanations, helping educators validate the accuracy of the questions.For example, in this "Beginner Level" question, students are asked to identify the correct syntax of an if statement in the C programming language. ChatGPT provides answers with the explanation as shown in Figure \ref{fig:educator_question_easy-complex} (a). 

Educators save time on creating assessments, as ChatGPT can generate relevant questions quickly, helping streamline lesson planning and evaluation processes.
Overall, ChatGPT enhances the creation of tailored, high-quality educational assessments, freeing educators to focus more on teaching and engagement. ChatGPT also brings the ability to make efficient quizzes and assignment creation, as well as automated grading and feedback for any specific topics in the engineering domain, among others. Assessment is a critical component of the educational process, but creating quizzes and assignments that accurately measure student understanding can be labour-intensive. ChatGPT can streamline this process by generating assessments tailored to specific topics and difficulty levels:
\begin{itemize}
    \item Example: An educator teaching a course on data structures might ask ChatGPT to "Create a set of challenging questions on binary trees and graph theory." ChatGPT could then generate a quiz that includes both multiple-choice questions and coding exercises designed to test a student’s comprehension and problem-solving skills. Figure \ref{fig:educator_question_easy-complex} (b) shows the question-setting ability of the ChatGPT for educators. 
\end{itemize}

ChatGPT can also assist with grading assignments and providing feedback. For instance, after students complete a writing assignment, ChatGPT could be used to provide initial feedback on grammar, sentence structure, and content coherence. In subjects like mathematics or programming, ChatGPT could even automate the grading of assignments, ensuring accuracy and consistency while freeing up valuable time for educators to focus on interactive and creative teaching activities.
\subsubsection{Teaching Materials and Slide Preparation Support}
\begin{figure}
    \centering
    \includegraphics[width=0.8\linewidth]{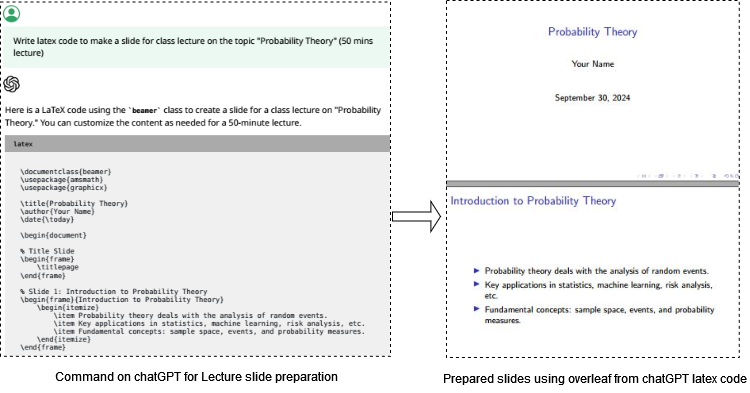}
    \caption{Lecture slide prepation with chatGPT and Overleaf}
    \label{fig:Lectureslide}
\end{figure}

ChatGPT enhances essential writing and communication skills in engineering by providing real-time feedback on grammar, vocabulary, and phrasing. Educators can use ChatGPT to review student theses, translate reports, and offer constructive comments to improve manuscripts. Additionally, ChatGPT supports language learning through interactive exercises, allowing students to practice conversational skills via simulated dialogues. In multilingual classrooms, ChatGPT assists with translating educational materials, ensuring that all students have access to resources in their preferred language, which is especially valuable in diverse settings. ChatGPT enables educators to create more inclusive, effective learning environments and supports researchers in accelerating their work. In combination with Overleaf, ChatGPT streamlines the creation of lecture slides. ChatGPT provides structured content, ideas, and sample LaTeX code, while Overleaf’s collaborative editor supports professional-quality slide design, particularly for math-focused subjects. This partnership improves slide content and allows educators to focus more on teaching.
The procedure has been shown in figure~\ref{fig:Lectureslide}.
\begin{figure}
    \centering
    \includegraphics[width=0.60\linewidth]{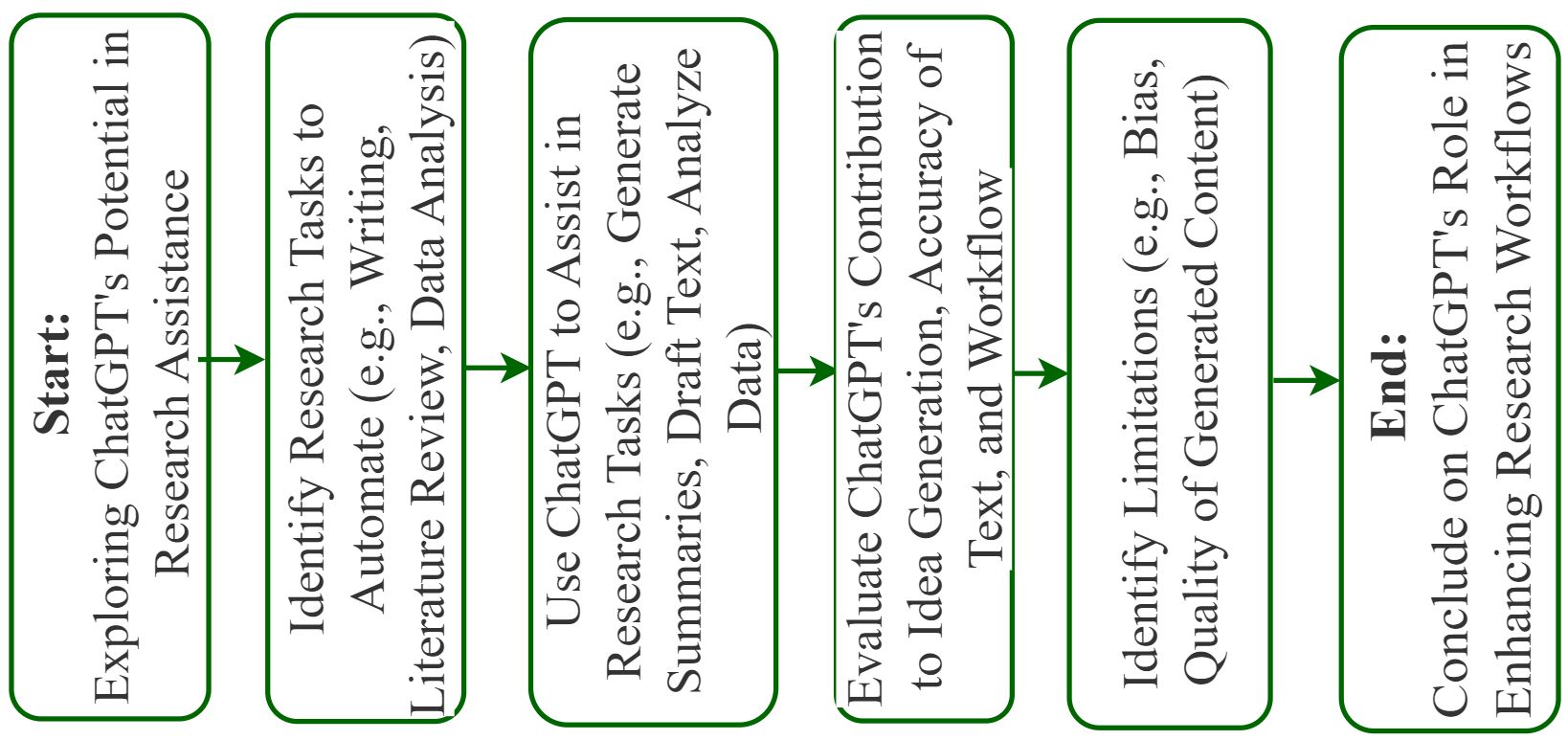}
    \caption{Opportunities and challenges for Researchers with ChatGPT}
    \label{fig:ChatGPT_Researchers}
\end{figure}

\subsection{Opportunities for Researchers} 
ChatGPT brings researchers many opportunities, significantly enhancing the research process from idea generation to publication and system deployment \cite{rahman2023chatgpt,miah2017motor,Mamun2024,miah2019eeg,Bantraffic_miah,zobaed2020real_miah, ali2022potential_miah,hossain2023crime_miah,rahim2024enhanced_miah,miah2019motor,kibria2020creation_miah,hossain2023exploring_miah,rahman2020comparative_miah,tusher2024}.
 This tool can assist in various stages of research, providing valuable support that allows researchers to focus more on the substance of their work. The detailed process is illustrated in Figure ~\ref{fig:ChatGPT_Researchers}.
   
\subsubsection{Writing Assistance Including Existing Research with Pros and Cons}
\begin{figure}[h]
\centering
\includegraphics[width=0.9\textwidth]{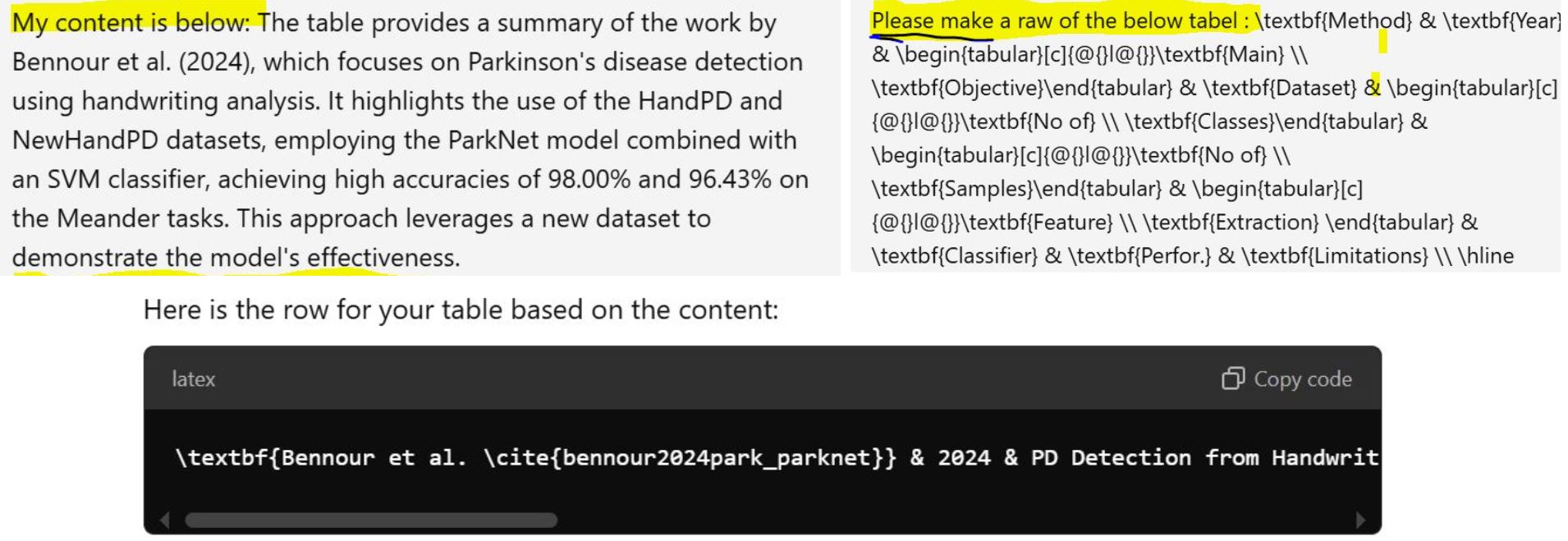}
\caption{Making a table raw from content}\label{fig:contntent-to-table}
\end{figure}

\begin{figure}[h]
\centering
\includegraphics[width=1.2\textwidth]{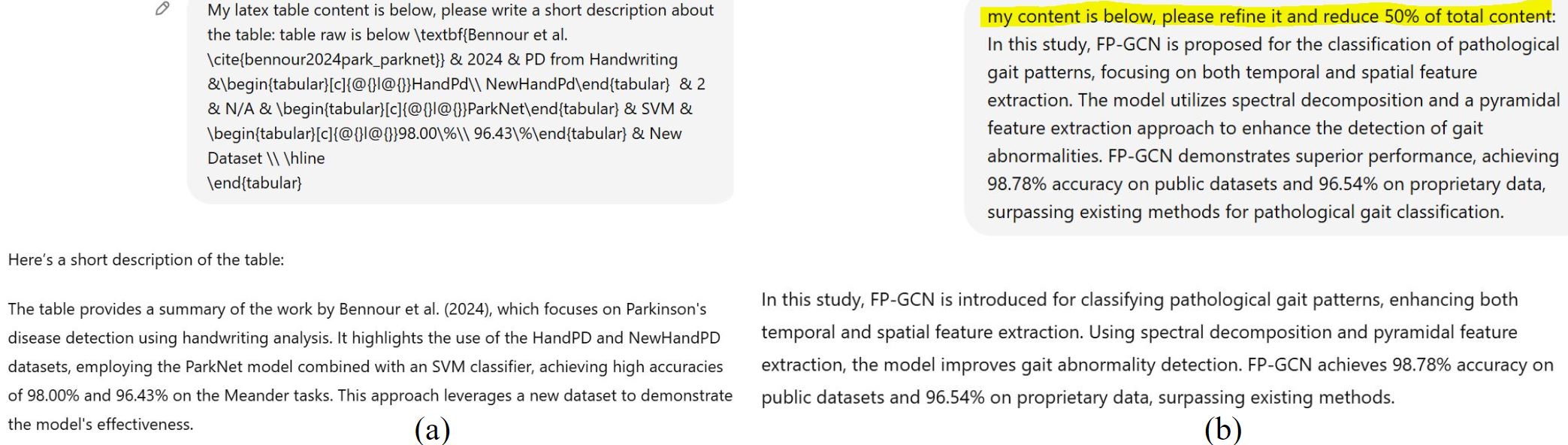}
\caption{ (a) Making content from table (b) Refine the content to reduce the words}\label{fig:table-content}
\end{figure}

ChatGPT can help researchers polish and improve their manuscripts by identifying and correcting typographical errors, resolving grammatical inconsistencies, and suggesting advanced vocabulary. We can also get help from ChatGPT to convert the manuscript into a latex version.  
  \begin{itemize}
       \item Example: Suppose we need to make a table from the existing work; in this case, we can give some information to the ChatGPT about an article and ask to make a table raw. ChatGPT can produce a table raw from the text; figure \ref{fig:contntent-to-table} shows the example like it.  In addition, researchers can take help to write the content about the table like Figure \ref{fig:table-content} (a). In addition, the researcher can refine their content using ChatGPT based on the requirement shown in Figure \ref{fig:table-content} (b). 
      \item Example: Suppose a researcher is writing a paper on the environmental impact of plastic waste. ChatGPT can help refine the language, ensuring the paper is well-structured and the arguments are presented logically, freeing the researcher to concentrate on the scientific content.
      
  \end{itemize}
  
  \begin{figure}[h]
\centering
\includegraphics[width=1.3\textwidth]{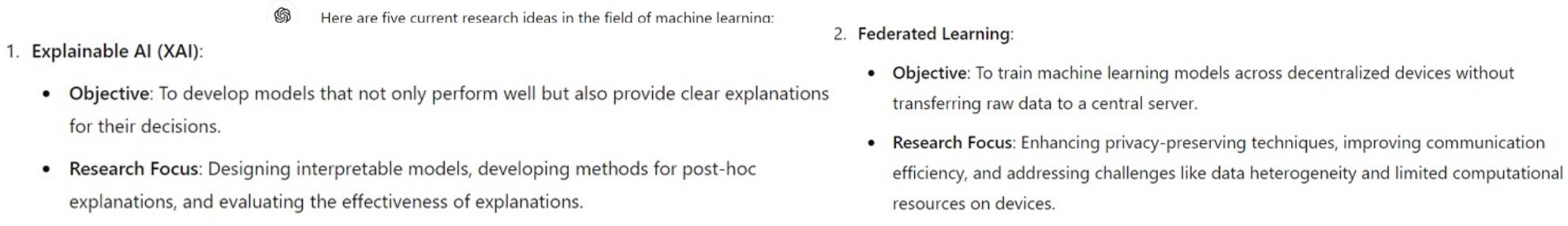}
\caption{Current research trends queries response}\label{fig:research_trends}
\end{figure}
    
ChatGPT can assist with literature reviews by summarizing existing research on a particular topic, specifically engineering topics, due to numerous difficulties in manual work. It can quickly sift through vast amounts of literature and provide concise summaries, highlighting key findings and identifying gaps that could be explored further. For instance, a researcher focusing on renewable energy could ask ChatGPT to "Summarize the latest research on solar panel efficiency." The model would then provide a summary of recent studies, pointing out trends and areas where further research is needed.
ChatGPT can give the latest updated work list and their problem, which helps researchers to come up with new ideas to solve the recent issue. One of the main problems of researchers in any domain is to find the update latest work on the corresponding field and its drawbacks with solvable clues or biomarkers. These biomarkers help researchers find immediate solutions that can help the world improve. This can inspire researchers to explore new directions in their work. For instance, if a researcher asks ChatGPT for an unexplored topic related to "reducing errors in time-constrained programming environments," the model might propose developing adaptive algorithms that dynamically adjust resource allocation based on real-time performance metrics. This could lead to a fresh line of inquiry that the researcher had not previously considered.

 \subsubsection{Data Analysis Support with Designing Flowchart}
 ChatGPT can assist researchers in selecting the appropriate statistical methods for their data analysis. It can explain various statistical techniques and recommend the best methods based on the research questions and the nature of the data. For instance, a survey researcher might be uncertain about which statistical tests to use. By asking ChatGPT for guidance, they could receive suggestions on the most suitable tests, such as a chi-square test for categorical data or a t-test for comparing means.

\begin{figure}
     \centering
     \includegraphics[width=0.8\linewidth]{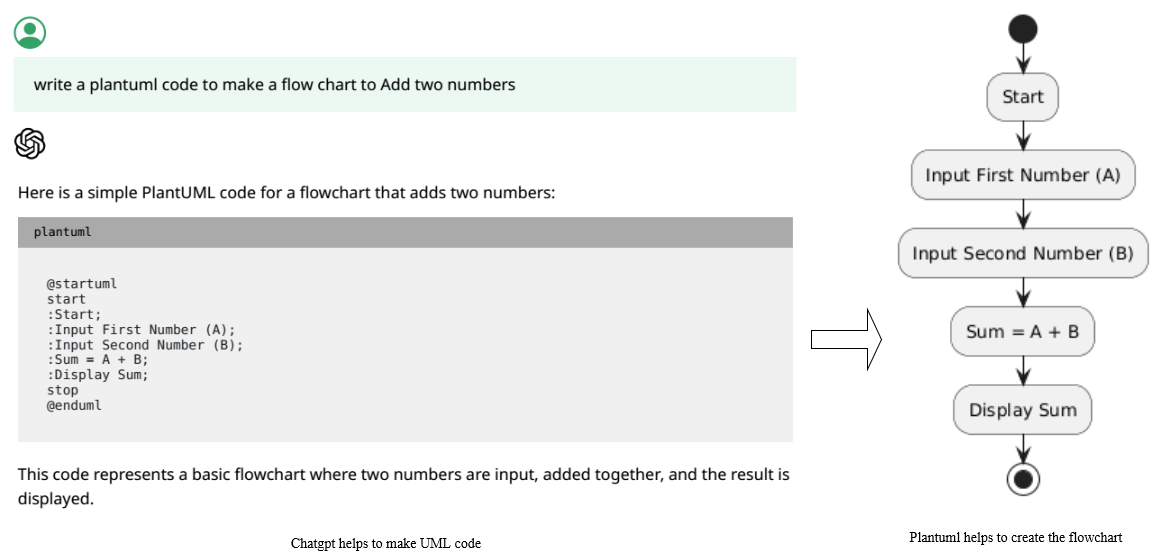}
     \caption{Chatgpt helps to design flow chart}
     \label{fig:Chatgpt in flow chart design}
 \end{figure}

ChatGPT is a versatile tool that can significantly enhance the research process by offering support in writing, literature review, idea generation, and data analysis. Its ability to engage in natural language conversations makes it an accessible and valuable resource for researchers across various disciplines. By leveraging ChatGPT's capabilities, researchers can streamline their workflows, generate new ideas, and ultimately produce higher-quality research.
ChatGPT can assist in designing flowcharts by providing step-by-step guidance and suggestions for visualizing processes, workflows, or algorithms. Describing the logic or sequence of actions to ChatGPT can suggest how to organize steps in a flowchart, identify decision points, and clarify the flow between tasks. Additionally, it can offer ideas for optimizing the structure and logic of the chart, making it more efficient and easy to follow. Moreover, ChatGPT can provide PlantUML code, which can be visualized using the PlantUML website, enabling users to create and view professional flowcharts in a simple, text-based format. This combination streamlines the flowchart design process, making it more accessible and customizable. The process of making flow chart using chatgpt and plantuml website has been shown on figure~\ref{fig:Chatgpt in flow chart design}.

 \subsubsection{ChatGPT's Role as a Research Guidance and Structuring Tool}
\begin{figure} [H]
     \centering
    \includegraphics[width=1.2\linewidth]
   {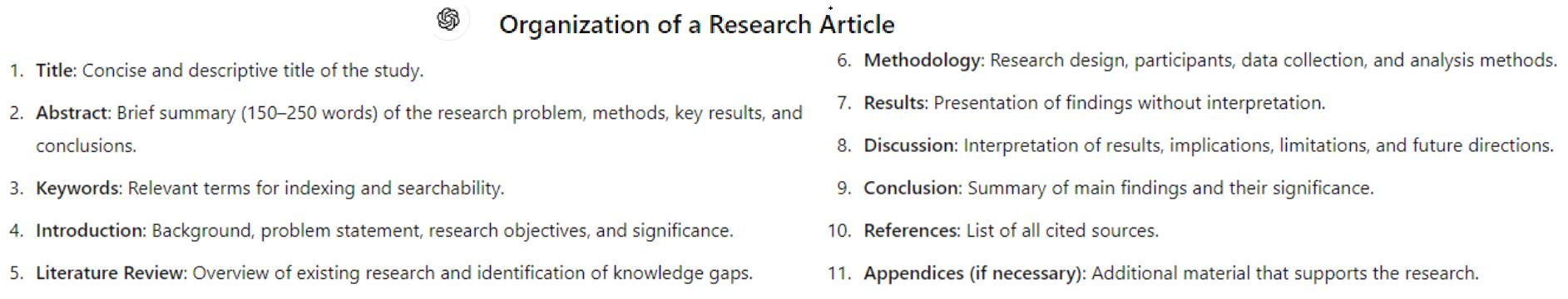}
     \caption{ChatGPT's Response to the Question: Organization of a Research Article}
     \label{fig:organization_of_a_research_article}
 \end{figure}

 ChatGPT is a valuable tool for researchers, assisting with the development, organization, and presentation of research work. It provides guidance on research methodologies, academic writing standards, and paper structure, enhancing clarity and cohesion.
For instance, ChatGPT offers specific support for beginner researchers, such as crafting focused titles, developing concise abstracts, and suggesting content for each section. It helps outline the introduction with relevant background, problem statements, and objectives and identifies key sources and knowledge gaps for the literature review. In the methodology, ChatGPT aids in describing research design, data collection, and analysis. It also advises on objective data presentation in the results section and interpretation in the discussion. By structuring content effectively, ChatGPT enables researchers to communicate ideas clearly.
Figure ~\ref{fig:organization_of_a_research_article} shows ChatGPT's Response to the Question: Organization of a Research Article.

Moreover, When writing a research paper, ChatGPT can clarify the difference between commonly confused sections, like the abstract and conclusion. It provides researchers with concise explanations, ensuring they correctly format and structure these sections:

\begin{itemize}[left=2cm] 
    \item \textbf{Abstract:} A brief summary of the research, including the problem, methodology, key results, and significance.
    \item \textbf{Conclusion:} A reflection on the results, discussing their implications, limitations, and future directions.
\end{itemize}

\begin{figure}[H]
     \centering
     \includegraphics[width=1.2\linewidth]{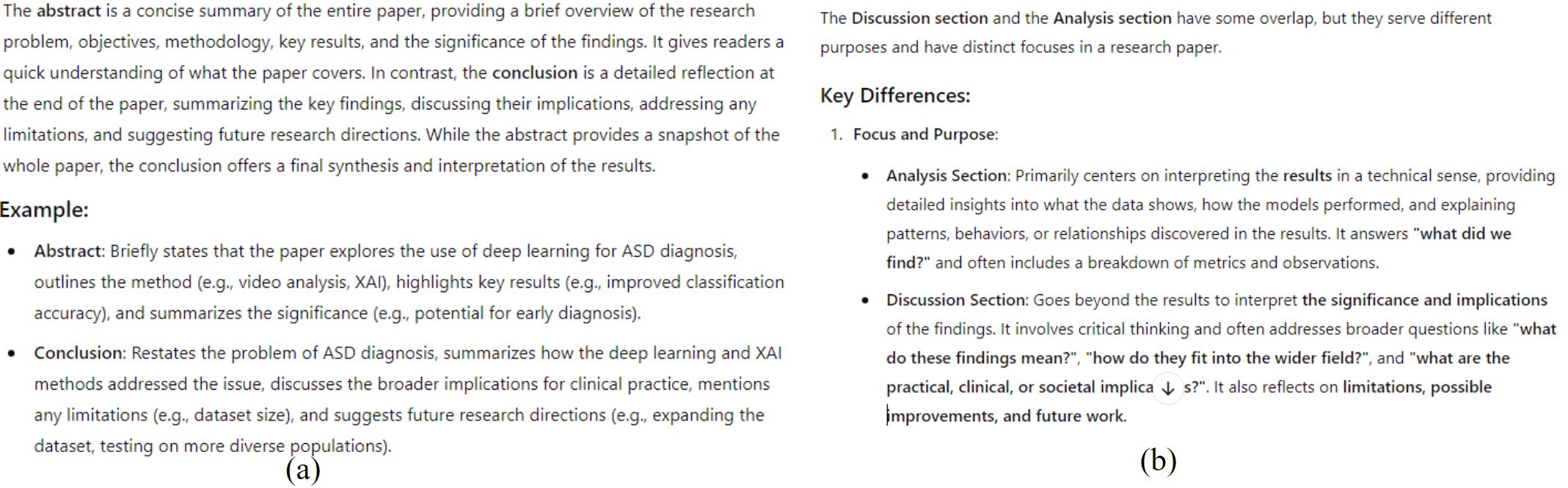}
     \caption{ChatGPT's Response to the Question: (a) (b)Analysis Section vs. Discussion Section}
     \label{fig:abstractvscon_analysis_and_discussion}
 \end{figure}

Figure ~\ref{fig:abstractvscon_analysis_and_discussion} (a) shows ChatGPT's Response to the Question: Abstract vs. Conclusion.
The Discussion and Analysis sections in a research paper often overlap, but each serves a unique purpose and has a distinct focus. This can sometimes cause confusion for researchers when organizing content under these headings. ChatGPT can assist by guiding researchers on how to structure and differentiate the content in such cases, as shown in Figure  ~\ref{fig:abstractvscon_analysis_and_discussion}(b).

This type of guidance helps researchers avoid common pitfalls and improve the quality and structure of their academic writing.
 
 \subsubsection{Assist to write Latex Codes for Paper}
ChatGPT serves as an effective tool for writing LaTeX code when preparing academic papers. It assists users in formatting documents according to specific journal or conference guidelines, offering code snippets for sections like title pages, abstracts, citations, references, figures, tables, and equations. ChatGPT also recommends LaTeX packages to enhance functionality and aesthetics, such as managing complex layouts, cross-referencing, and handling bibliographies with BibTeX. From structuring the paper to debugging errors, ChatGPT streamlines the LaTeX process, saving time and minimizing common frustrations. Using ChatGPT with Overleaf, users can quickly create LaTeX templates tailored to IEEE or Springer formats by requesting sample code with commands like, “Write a sample LaTeX code for IEEE or Springer conference paper format.”
  
\subsection{Programming Learning with ChatGPT}
\begin{figure}
    \centering
    \includegraphics[width=0.60\linewidth]{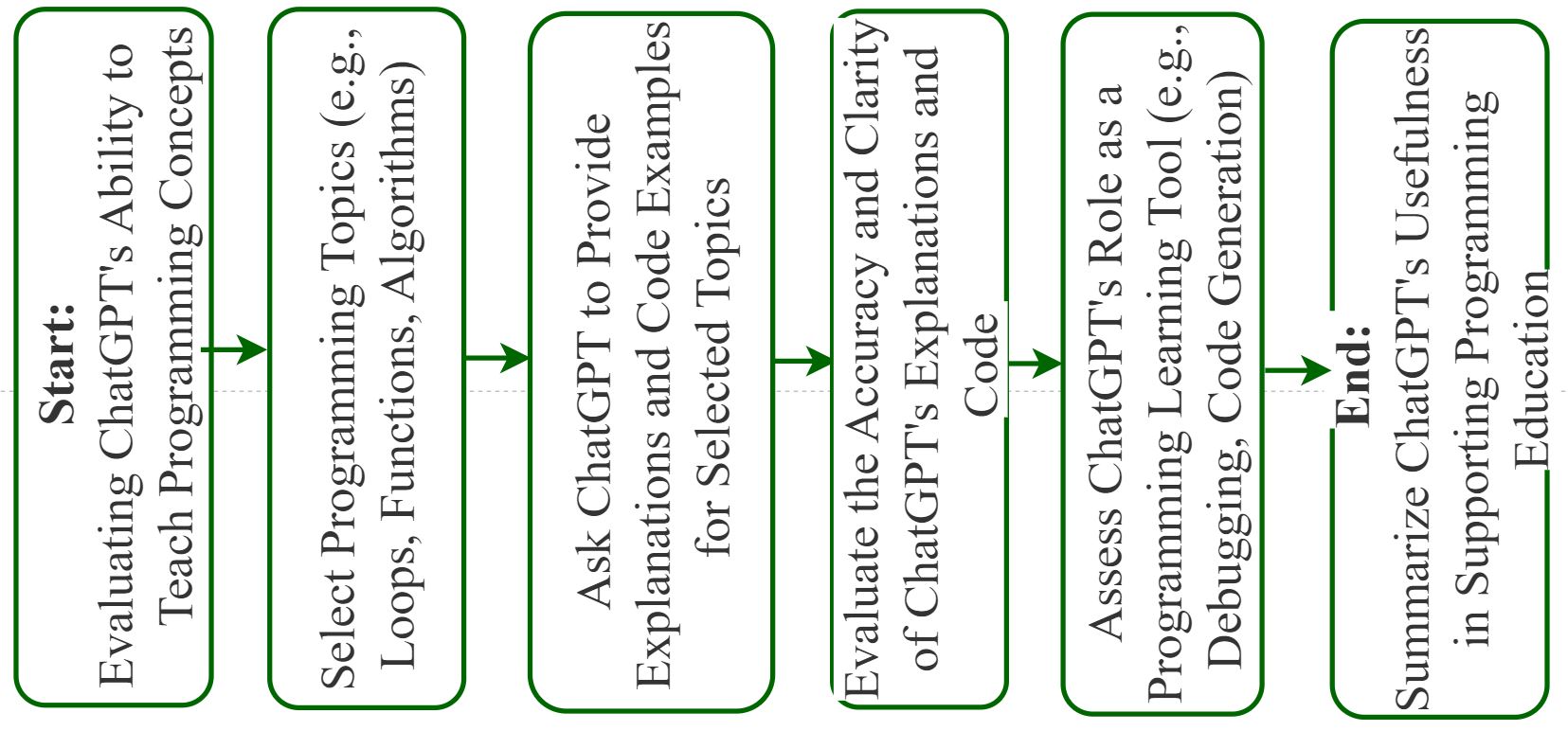}
    \caption{Opportunities and challenges for Programmer with ChatGPT}
    \label{fig:ChatGPT_Programmer}
\end{figure}

In the rapidly evolving field of computer science, programming is crucial for both academic and professional success. Mastery of programming languages and concepts requires regular practice and a strong conceptual foundation. ChatGPT, with its Transformer-based architecture, provides valuable support for programming education through code generation, error detection, and optimization. The detailed process is illustrated in Figure~\ref{fig:ChatGPT_Programmer}.
\subsubsection{Conceptual Understanding} 
Mastering programming requires a strong grasp of core concepts like variables, loops, functions, data structures, and algorithms. ChatGPT effectively breaks down these topics into easily understood explanations tailored to the learner's level. For example, if a beginner is struggling with a "for loop," ChatGPT can offer a simple explanation along with a basic Python example.
ChatGPT also handles advanced topics like object-oriented programming (OOP), recursion, and dynamic programming. For instance, when asked about "polymorphism" in OOP, ChatGPT explains how it allows objects of different classes to be treated as objects of a common superclass, with appropriate methods called based on the actual class at runtime.
Additionally, ChatGPT aids in understanding algorithms by providing pseudocode and a step-by-step breakdown, which helps learners visualize the process before coding. For example, in "merge sort," ChatGPT can generate pseudocode and explain the divide-and-conquer strategy used in the algorithm.

\subsubsection{Solution Code Generation} 
ChatGPT can generate complete code solutions based on a problem description, making it particularly helpful for learners needing guidance on coding approaches. It can produce code in languages like Python, Java, C++, and more. For example, if a learner is tasked with creating a program to calculate the Fibonacci sequence, they can ask ChatGPT to generate the code.
ChatGPT also adjusts code complexity based on the user’s level. For beginners, it might suggest a simple iterative solution; for advanced users, it could provide a recursive approach or introduce memoization for optimization. This feature is invaluable in educational settings, especially in competitive programming, where learners must implement algorithms efficiently. ChatGPT can quickly generate solutions that learners can analyze and refine to deepen their understanding and coding skills. 

\subsubsection{Error Detection and Optimization}

\begin{figure}[h]
\centering
\includegraphics[width=0.99\textwidth]{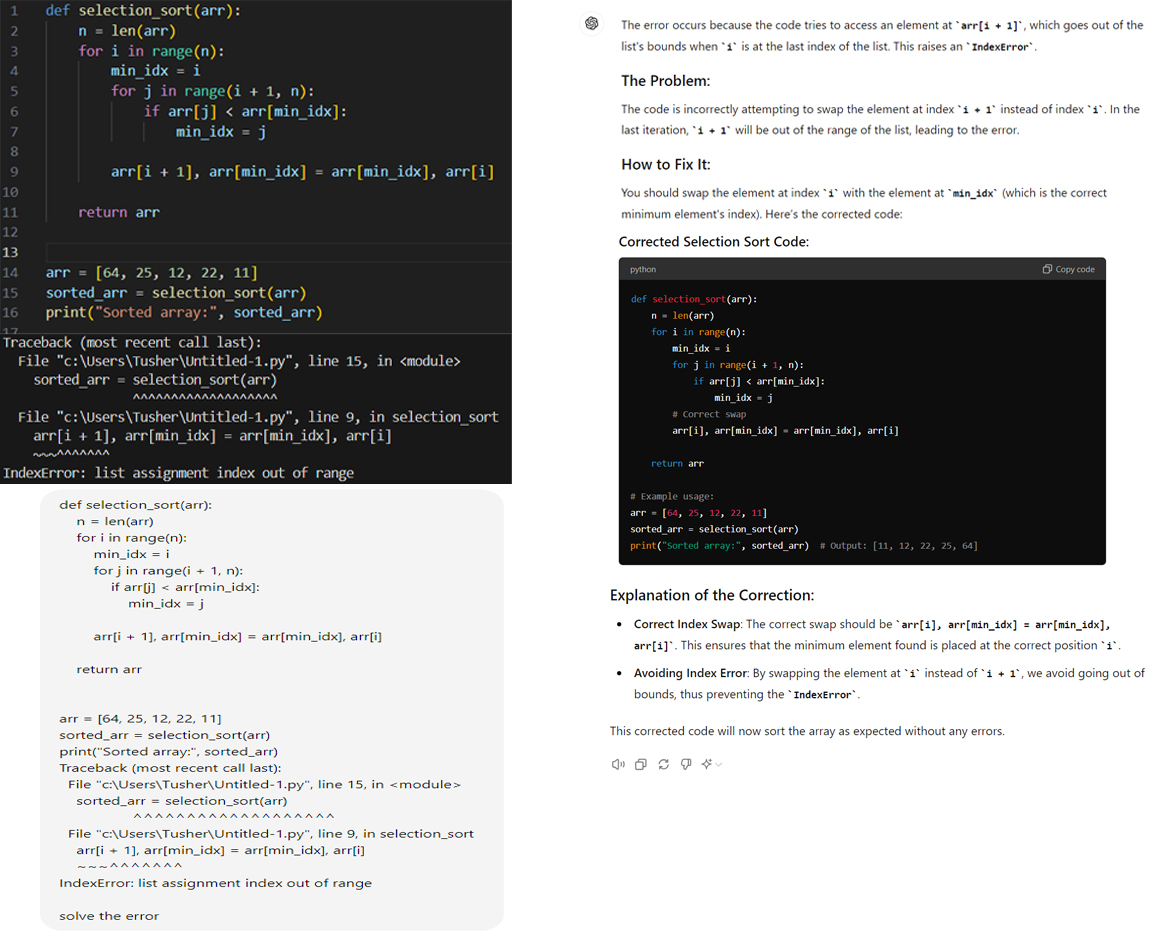}
\caption{Query: Error detection}\label{fig:edu-error-detection}
\end{figure}

Debugging and identifying errors is one of the toughest aspects of programming. ChatGPT can assist learners by spotting both syntax and logical errors in their code. Analyzing the provided code can pinpoint issues and suggest corrections. For example, if a student encounters a "TypeError" in their Python function, they can paste the code into ChatGPT along with a description of the error. ChatGPT would then analyze the code, identify the root cause, and recommend a fix, such as adjusting the data type or modifying a function cal. Figure \ref{fig:edu-error-detection} shows the error detection from a program. 

   ChatGPT provides valuable guidance on coding best practices, such as using clear variable names, commenting code, and following style guides like PEP 8 in Python. It also suggests optimizations to improve performance, such as reducing time complexity and memory usage.
For example, if a student uses bubble sort $$(O(n^2))$$ to sort an array, ChatGPT might recommend quicksort, with a more efficient average time complexity of $O(n log n)$.
As learners work on coding problems, they can refine solutions with ChatGPT’s real-time feedback, improving both code quality and their understanding of key concepts. ChatGPT supports all aspects of programming education, from foundational learning to advanced problem-solving, by explaining concepts, generating code, detecting errors, and suggesting improvements. As a tutor, code generator, and debugging assistant, ChatGPT makes programming more accessible and engaging for learners at all levels.

\section{Methodology of Analysis with Our Newly Created Dataset}\label{sec2}
The study aimed to evaluate ChatGPT's effectiveness in solving subject-specific problems, its reliability in providing educational support, and its perceived value among users. A comprehensive experimental framework was designed, which included subjective problem-solving experiments across subjects like mathematics, programming, and electronics, where ChatGPT's solutions were compared with traditional textbook answers. Surveys were distributed to students and teachers to assess ChatGPT's impact on learning, research assistance, and programming education, measuring perceived learning improvement, ease of understanding, and information reliability.

Data was collected from engineering students and teachers through surveys with 15 questions targeting ChatGPT’s role in research, problem-solving, programming, and essay writing. Feedback from students, teachers, and researchers was gathered on ChatGPT’s ability to enhance understanding, aid in lesson planning, and support research tasks. Statistical analysis, including descriptive and thematic approaches, was used to interpret the data and provide insights into the impact of ChatGPT in education and research.

\subsection{Study Design}
This study utilized a quantitative survey approach to explore the impact of ChatGPT on the academic experiences of engineering students and teachers. The survey, hosted on Google Forms, included 15 questions designed to capture various aspects of ChatGPT usage, particularly in academic tasks like research, problem-solving, programming, and essay writing. The survey's objective was to assess the perceived benefits, challenges, and overall effectiveness of ChatGPT in an educational context. The survey was conducted over two weeks, gathering responses from participants across different academic stages.

\subsection{Sample}
The sample consisted of 81 participants from engineering backgrounds, comprising both students and teachers. The student respondents were categorized based on their academic standing, ensuring a diverse pool of experiences from different stages in their academic journey. The distribution of students was as follows: 18.5\% from the 1st semester/year, 27.2\% from the 2nd year, 14.8\% from the 3rd year, 23.5\% from the 4th year, and 16\% of respondents who had completed their studies. This wide representation ensured that the study captured a holistic view of ChatGPT’s role across varying levels of experience with academic challenges.
\begin{table}[ht]
\centering
\caption{Survey on how ChatGPT helps students.} \label{tab:survey_output_yes_no_every_question}
\begin{tabular}{|c|p{8cm}|c|c|}
\hline
\textbf{Number} & \textbf{Question} & \textbf{Yes (\%)} & \textbf{No (\%)} \\ 
\hline
Q1 & Have you used ChatGPT to get quick answers to your academic questions? & 91.40\% & 8.60\% \\ 
\hline
Q2 & Has ChatGPT helped you find reliable sources for your research projects? & 79\% & 21\% \\ 
\hline
Q3 & Have you used ChatGPT to solve complex mathematical equations? & 56.00\% & 44.00\% \\ 
\hline
Q4 & Have you used ChatGPT to generate ideas for essays or writing assignments? & 84\% & 16\% \\ 
\hline
Q5 & Have you used ChatGPT to understand complex scientific concepts in subjects like physics or chemistry? & 53.10\% & 46.90\% \\ 
\hline
Q6 & Has ChatGPT helped you complete your lab reports or scientific research? & 75.30\% & 24.70\% \\ 
\hline
Q7 & Have you used ChatGPT to debug your programming code? & 85.20\% & 14.80\% \\ 
\hline
Q8 & Have you taken help/support from ChatGPT for solving programming problems? & 87.70\% & 12.30\% \\ 
\hline
Q9 & Are you using paid ChatGPT? & 5\% & 95\% \\ 
\hline
\end{tabular}

\end{table}
\subsection{Data Collection}
\begin{figure}[h]
\centering
\includegraphics[width=0.8\textwidth]{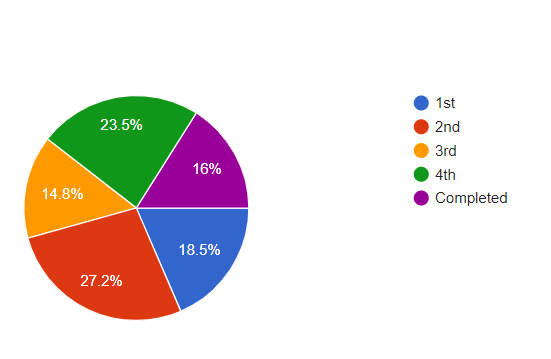}
\caption{Year-wise engineering student who participated in our data collection procedure. \label{fig:year-wise-student}}
\end{figure}

The data were collected via a 15-question survey administered through Google Forms. The questions covered a range of topics, including the frequency of ChatGPT use, its role in solving academic problems, and specific tasks for which it was most useful. We collected the dataset from the engineering background students and teachers. Figure \ref{fig:year-wise-student} shows the year-wise student ratio who participated in our data collection procedure.  According to the figure, 18.50\% are students, 27.2\% are the 2nd year students, 14.80\% are the 3rd year students, and 23.5\% students join from 4th year students. In addition, graduates and teachers participated in the data collection procedure, and 16\% of the team participated. The data collection survey included:
\begin{itemize}
    \item 
Post-interaction feedback was gathered to evaluate how ChatGPT enhanced understanding, problem-solving skills, and its perceived value as a learning tool. Likert scale questions were also used to gauge attitudes toward ChatGPT's usefulness in academic tasks.
    \item 
      Teachers were asked to evaluate how ChatGPT aided in lesson planning, question answering, and explaining complex topics.
    \item Researchers provided insights on ChatGPT’s ability to assist in literature reviews, idea generation, and overall research support.
    \item
    Multiple-choice questions focused on identifying areas where ChatGPT provided the most value, such as research assistance, programming support, or generating ideas for assignments.
     \item 
    Open-ended questions will be used to gather qualitative insights on participants' suggestions for improving ChatGPT.
\end{itemize}

\subsection{Statistical Snslydid of Newly Collected Data}
The data were analyzed using both quantitative and qualitative methods. Descriptive statistics, such as percentages and frequency distributions, summarized responses from the Likert scale and multiple-choice questions. Comparative analysis highlighted differences between students and teachers in their use of ChatGPT. Thematic analysis of open-ended responses identified key trends and suggestions for improvement. Graphs and tables were used to represent the findings for easier visual interpretation.

\subsection{Results}
After analyzing the newly collected dataset, we generated several statistical outputs in ratios and visualization figures, which are presented below.
\subsubsection{General and  Research-Based Findings }
Table \ref{tab:survey_output_yes_no_every_question} shows he general finding of the data analysis. The survey revealed that 91.4\% of respondents had used ChatGPT to get quick answers to academic questions, demonstrating its widespread adoption as a tool for immediate problem-solving. The largest group of users came from 2nd-year students (27.2\%), suggesting that early-stage university students are particularly likely to explore AI tools like ChatGPT for academic help. Only 8.6\% of participants indicated that they had not used ChatGPT for academic queries, underscoring its role as a go-to resource.

79\% of respondents found ChatGPT helpful for sourcing research materials, while 21\% experienced difficulties with the reliability of the sources provided. Additionally, 84\% of students reported using ChatGPT to generate ideas for essays or writing assignments, indicating its significant role in assisting with brainstorming and overcoming writer’s block. However, feedback highlighted a need for more accurate and reliable references, with some students reporting that ChatGPT occasionally produced fake or incomplete citations.

\subsubsection{Use of ChatGPT in Solving Academic and Programming Based Project Problems}

When asked about ChatGPT’s role in tackling complex academic tasks, 55.6\% of respondents had used it to solve mathematical equations, while 53.1\% found it helpful in understanding complex scientific concepts. These results suggest a mixed level of success, with nearly half of the participants noting that ChatGPT did not fully meet their needs in more technical subjects. The respondents suggested that ChatGPT’s ability to handle advanced math and science problems required improvement.
\begin{figure}[h]
\centering
\includegraphics[width=0.8\textwidth]{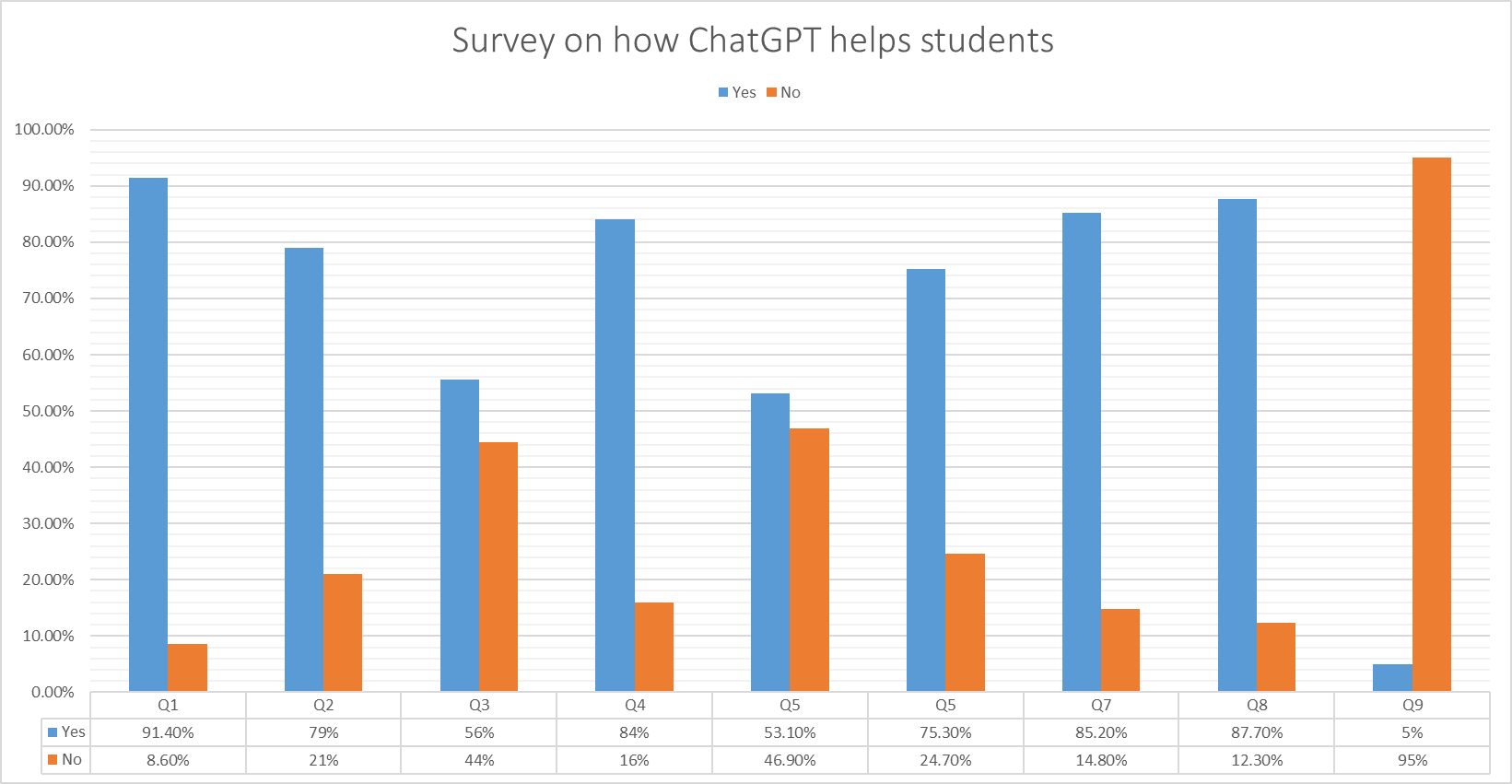}
\caption{Survey on how ChatGPT helps students. Q1-Q9 refer in Table \ref{tab:survey_output_yes_no_every_question} }\label{fig:survey}
\end{figure}

The survey highlighted ChatGPT’s strong performance in aiding programming tasks, with 85.2\% of respondents using it for debugging code and 87.7\% seeking support for solving programming-related problems. These high response rates demonstrate ChatGPT’s effectiveness in assisting with coding challenges, a critical area of support for engineering students.

\subsubsection{Free vs. Paid Versions}

One of the more surprising results came from the question regarding the use of paid ChatGPT versions. Only 5\% of respondents were using the paid model, while 95\% relied on the free version. This suggests that most students find the free version adequate for their needs or that financial constraints deter them from upgrading to the paid model.

\subsubsection{Feedback and Suggestions for Improvement}

The open-ended responses provided several key suggestions for improving ChatGPT, including better handling of complex mathematical problems, enhanced image and document recognition features, and the addition of voice interaction capabilities. A notable number of students expressed frustration with ChatGPT providing incorrect answers, particularly in programming tasks and advanced problem-solving.

\section{Discussion About Findings, Threats and Strategies}\label{sec12}
ChatGPT offers valuable capabilities in education and research but also presents challenges, especially in technical fields like programming. Its ability to generate human-like content raises concerns about appropriate use in these contexts. This section discusses key findings, implications for education, challenges of using ChatGPT, and potential strategies to address these issues.

\subsection{Interpretation of Findings and Implications for Education}
The results show that students widely use and appreciate ChatGPT, particularly for quick answers, programming help, and idea generation. However, improvements are needed in providing accurate references and handling complex math and science queries. With only 5\% of students using the paid version, it suggests that while the free version meets basic needs, financial barriers may limit access to paid features.
These findings suggest that AI tools like ChatGPT have the potential to significantly enhance academic support, especially in fields like engineering. Institutions may consider integrating ChatGPT into their academic support services, particularly for programming and writing assistance. However, improvements in the tool’s accuracy and reliability will be necessary to ensure it remains a trustworthy resource for students tackling complex tasks.

\subsection{Integrity of Assignments and Online Exams: The Impact on General and Engineering Students}
The rise of online exams has heightened concerns about academic integrity, especially with AI tools like ChatGPT. Engineering students, familiar with AI and programming, can leverage ChatGPT for advanced tasks like coding and algorithm solving, making misuse harder to detect, particularly in developing countries with limited resources \cite{Susnjak2022ChatGPT}.
To address these challenges, institutions should create assessments that emphasize critical thinking and problem-solving skills that AI cannot easily replicate. Clear ethical guidelines, advanced plagiarism detection, stronger online proctoring, and requiring drafts or oral defenses can further ensure integrity, particularly in engineering fields. Continued research is essential to understand AI’s impact and to maintain academic standards \cite{cotton2023chatting}.

\subsection{Blind Reliance on Generative AI Tools and Its  Difficulty in Evaluating Generated Answers}
Overreliance on generative AI tools like ChatGPT can negatively impact education and research by potentially weakening critical thinking and problem-solving skills. While ChatGPT offers quick solutions for problem-solving and text generation, its use in essays and research raises concerns \cite{Stokel-Walker2023ChatGPT}. Sam Altman, CEO of OpenAI, cautions that ChatGPT is “limited” and not yet reliable for crucial tasks, emphasizing the need for greater robustness and truthfulness.
Students, educators, and researchers should recognize these limitations and use AI to support, not replace, genuine understanding and original thinking \cite{Pavlik2023Collaborating}. Developing curricula and assessments that account for AI use can help ensure that educational practices still foster essential skills.
As an AI model, ChatGPT uses algorithms to generate responses based on learned patterns, making its output increasingly similar to human-generated text. This similarity challenges educators and researchers as current plagiarism tools struggle to distinguish AI-generated content \cite{cotton2023chatting, Elkins2020Can, Gao2022Comparing, Dehouche2021Plagiarism}. Some institutions are now restricting ChatGPT use \cite{Kalhan2023ChatGPT}. Cotton et al. \cite{cotton2023chatting} suggest identifying AI-generated texts by noting inconsistencies, lack of citations, factual errors, and ambiguous language. Further research is needed to develop AI-based plagiarism detection tools to uphold integrity in education and research.

\subsection{Critical Thinking and Problem-Solving Skills}
Generative AI tools like ChatGPT can provide accurate or partially correct answers to technical questions and generate programming code based on problem descriptions and algorithms. While these features offer quick access to information, over-reliance on ChatGPT can hinder the development of critical thinking and problem-solving skills, especially in engineering education. Engineering students must cultivate strong analytical and creative problem-solving abilities across fields like mechanical, electrical, and civil engineering.
In engineering, solving complex problems requires deep understanding, critical analysis, and applying theoretical knowledge to real-world scenarios. Relying on ChatGPT for answers may prevent students from fully engaging with the material. Additionally, there are no reliable tools to detect AI-generated code or solutions, making it difficult to prevent misuse in exams, projects, and competitions.
To address these concerns, strategies must be developed to identify AI-generated responses, programming codes, or engineering solutions.

\begin{itemize}
    \item \textbf{Look for Telltale Signs:} Responses generated by ChatGPT often lack personalization and may appear overly generic. In the context of engineering, AI-generated solutions may also contain inconsistencies, over-simplifications, or fail to consider practical constraints that a human engineer would typically address.
    \item \textbf{Check for Coherence:} AI-generated responses may lack a consistent or logical flow, especially when dealing with complex, multidisciplinary engineering problems. Disjointed, incomplete, or nonsensical answers could indicate that they were generated by an AI model rather than developed through human reasoning.
    \item \textbf{Compare Responses:} Comparing AI-generated solutions with those created by students or other human experts can help identify if the content has been replicated. Identical or closely matching responses may suggest reliance on AI-generated content.
    \item \textbf{Use Plagiarism-Detection Tools:} Advanced plagiarism detection tools can be employed to check if an answer, engineering calculation, or design has been copied from other sources, including online repositories or AI-generated databases. These tools can be particularly useful in identifying cases where students submit AI-generated content as their own work.
    \item \textbf{Ask Follow-up Questions:} In instances where there is suspicion of AI-generated content, educators can ask follow-up questions that require students to explain their reasoning, the steps they took to arrive at a solution, or how they would adapt their approach to different scenarios. This can help assess the depth of the student's understanding and their ability to apply knowledge independently.
    \item \textbf{Design Thought-Provoking Assessments:} Educators can create assignments and exams that require students to apply critical thinking and creativity, such as open-ended design problems, case studies, or real-world scenarios that are less likely to have straightforward, AI-generated solutions.
    \item \textbf{Promote Ethical Awareness:} Institutions should emphasize the importance of academic integrity and the ethical implications of relying on AI-generated content. Encouraging a culture of honesty and original thinking can help mitigate the misuse of tools like ChatGPT.
\end{itemize}

By adopting these strategies, educators across all engineering disciplines can help ensure that students develop the critical thinking and problem-solving skills essential for their future careers while addressing the challenges posed by the increasing use of AI tools in education.

\subsection{Limitations of the ChatGPT Based on Our Analysis}
It is important to note that our experiments were conducted using ChatGPT, which is currently in an active development phase. During these experiments, we observed significant results in code generation, error checking and debugging, and solution code optimization. However, the results obtained may vary for several reasons: 
\begin{itemize}
    \item The study's limitation includes its sample size and the fact that it only surveyed students and teachers from an engineering background, which may limit the generalizability of the findings across other disciplines.
    \item The release of a new version of ChatGPT may lead to different outcomes.
    \item Asking different questions from those used in this study could produce varying results.
    \item The nature of the problem descriptions provided can impact the results.
    \item Code optimization results may differ depending on the initial solution codes.
\end{itemize}
Further research could include a more diverse sample and a larger population.
\section{Conclusion}\label{sec13}
ChatGPT and other AI language models (LLMs) have significant potential in education and research. ChatGPT can engage in human-like conversations and generate text nearly indistinguishable from human writing, making it useful for answering questions, writing essays, solving problems, explaining topics, tutoring, language practice, and supporting teaching and research. It addresses both technical (e.g., programming, engineering) and non-technical (e.g., language, literature) challenges.
Our surveys and experiments show that ChatGPT is valuable for programming and broader educational purposes. However, it has limitations, such as lacking common sense, potential biases, difficulties with complex reasoning, and an inability to process visual information. Users should be mindful of these shortcomings and avoid over-reliance on the tool.
Ethical concerns, including bias, privacy, security, misuse, accountability, and transparency, must be carefully considered when integrating AI into education and research. Despite these challenges, the risks associated with LLMs can be managed to ensure their reliable and equitable use in education and research.
\backmatter

\section*{Declarations}
\begin{itemize}
\item Funding
\item Conflict of interest/Competing interests (check journal-specific guidelines for which heading to use)
\item Ethics approval and consent to participate
\item Consent for publication
\item Data availability 
\item Materials availability
\item Code availability 
\item Author contribution
\end{itemize}
\noindent
If any of the sections are not relevant to your manuscript, please include the heading and write `Not applicable' for that section.

\bigskip

\bibliography{sn-bibliography}

\end{document}